\documentclass[12pt]{article}

\usepackage{newtxtext,newtxmath}

\usepackage{graphicx}

\usepackage[letterpaper,margin=1in]{geometry}

\renewenvironment{abstract}
	{\quotation}
	{\endquotation}

\date{}

\makeatletter
\renewcommand{\fnum@figure}{\textbf{Figure \thefigure}}
\renewcommand{\fnum@table}{\textbf{Table \thetable}}
\makeatother

\usepackage{scicite}

\usepackage{url}

\newcommand{\dbm}[1]{\dot{\boldsymbol{#1}}}
\newcommand{\ddbm}[1]{\ddot{\boldsymbol{#1}}}

\def\scititle{
	Agile and Cooperative Aerial Manipulation of a Cable-Suspended Load 
}
\title{\bfseries \boldmath \scititle}

\author{
	Sihao Sun$^{1\ast}$,
	Xuerui Wang$^{1}$,
	Dario Sanalitro$^{2}$,\and
        Antonio Franchi$^{3,4}$,
        Marco Tognon$^{5}$,
        Javier Alonso-Mora$^{1}$\and
	\small$^{1}$Delft University of Technology, the Netherlands.\and
    \small$^{2}$Sorbonne Université, France.\and
    \small$^{3}$University of Twente, the Netherlands.\and
    \small$^{4}$ Sapienza University of Rome, Italy \and
        \small$^{5}$Univ Rennes, CNRS, Inria, IRISA, France\and
        \small$^\ast$Corresponding author. Email: sihao.sun@outlook.com\and
}

\begin{document} 

\maketitle

\begin{abstract} \bfseries \boldmath
Quadrotors can carry slung loads to hard-to-reach locations at high speed.
Since a single quadrotor has limited payload capacities, using a team of quadrotors to collaboratively manipulate the full pose of a heavy object is a scalable and promising solution. 
However, existing control algorithms for multi-lifting systems only enable low-speed and low-acceleration operations due to the complex dynamic coupling between quadrotors and the load, limiting their use in time-critical missions such as search and rescue.
In this work, we present a solution to substantially enhance the agility of cable-suspended multi-lifting systems. 
Unlike traditional cascaded solutions, we introduce a trajectory-based framework that solves the whole-body kinodynamic motion planning problem online, accounting for the dynamic coupling effects and constraints between the quadrotors and the load. 
The planned trajectory is provided to the quadrotors as a reference in a receding-horizon fashion and is tracked by an onboard controller that observes and compensates for the cable tension. 
Real-world experiments demonstrate that our framework can achieve at least eight times greater acceleration than state-of-the-art methods to follow agile trajectories.
Our method can even perform complex maneuvers such as flying through narrow passages at high speed.
Additionally, it exhibits high robustness against load uncertainties, wind disturbances, and does not require adding any sensors to the load, demonstrating strong practicality.
\end{abstract}

\noindent
\section*{Video}
A video of the experiment can be found at \url{https://youtu.be/FBWN-rTK1YU}
\section*{Introduction}
Quadrotors stand out for their unparalleled agility, speed, and mobility compared to other robotic systems. This unique capability has made them highly suitable for lifting and transporting objects to hard-to-reach locations at high speed~\cite{hanover2021performance, saviolo2023learning}.
However, the payload capacity of a single quadrotor is limited, prompting the exploration of utilizing multiple quadrotors in collaboration to transport (position control) and even manipulate (full pose control) heavy objects, resulting in a multi-lifting system~\cite{loianno2017cooperative, tagliabue2017collaborative, tagliabue2019robust}.
This strategy has great potential in a wide range of applications requiring heavy object manipulation, such as construction, disaster relief, and agriculture, as well as space exploration missions on Mars and Titan, where aerial vehicles have very limited resources and payload capacity~\cite{balaram2021ingenuity,lorenz2018dragonfly}.
Among the various manipulation mechanisms, the cable-suspended solution stands out for its simplicity and low weight~\cite{foehn2017fast, panetsos2024gp, sreenath2013dynamics, li2021cooperative, li2023nonlinear, lee2013geometric, lee2017geometric, geng2022load, wahba2024efficient, wahba2023kinodynamic}.
By connecting each quadrotor to a different location on the load through cables, a team of three quadrotors, or more, can change the full pose of the cable-suspended load by adjusting their positions, eliminating the need for additional mechanisms like robotic manipulators.

However, existing cooperative autonomous flight algorithms can only achieve pose control of a cable-suspended object at low speed and low acceleration, greatly limiting its performance and endurance in time-critical missions.
The main challenge lies in addressing the complex dynamic coupling and kinematic constraints between the robots, cables, and the load.
Early works typically resort to a quasi-static assumption to neglect the dynamic coupling effects ~\cite{michael2011cooperative, manubens2013motion, fresk2013full, sanalitro2020full}, and only consider the kinematic constraints to determine the position and the path of quadrotors to reach the target pose of the load.
Despite being simple, failing to account for dynamic coupling leads to undesired swinging motions, and cannot guarantee a safe load distribution on each quadrotor.

\begin{figure}
    \centering
    \includegraphics[width=0.98\linewidth]{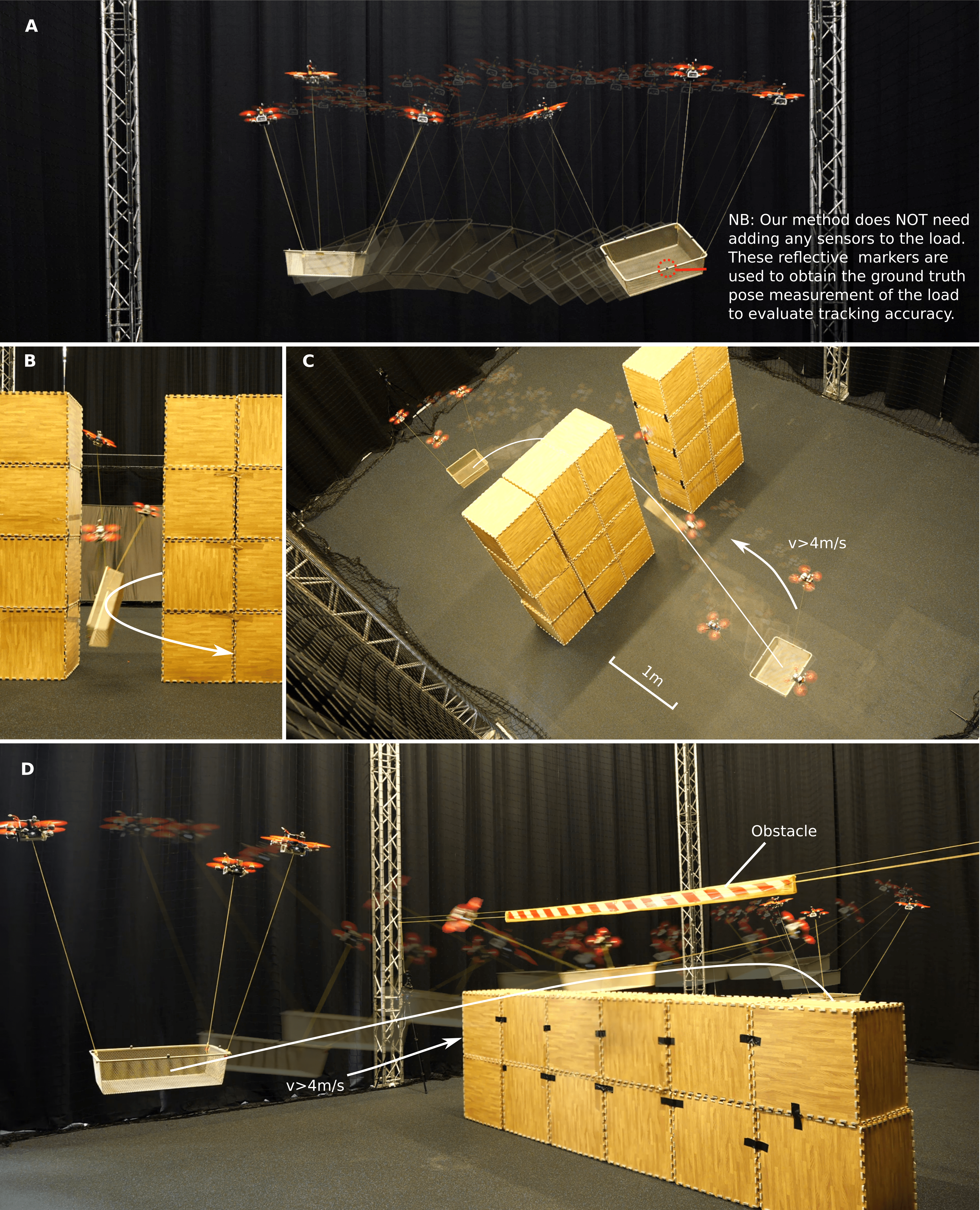}
    \caption{\textbf{Snapshot of the real-world experiments.} We propose an approach to control a cable-suspended load using multiple quadrotors with high agility. (\textbf{A}) Our approach enables agile full-pose control of a cable-suspended load. \textbf{(B-D)} It enables the quadrotors to dynamically control the load pose and fly through a narrow passage and a horizontally oriented gap. A summary of the experiments is highlighted in Movie 1.
    }
    \label{fig:eye_catcher}
\end{figure}

To account for dynamic coupling effects, recent works employ a force-based framework that employs the full dynamic model of the cable-suspended multi-lifting system.
Given the pose of the reference load, the methods in this framework calculate a desired wrench (force and torque) that acts on the load through an outer-loop controller, for example, inverse-dynamics control~\cite{masone2021shared}, nonlinear model-predictive control (NMPC)~\cite{li2023nonlinear,sun2023nonlinear}, and geometric control ~\cite{lee2017geometric, li2023rotortm, wahba2024efficient}.
Then the commanded wrench is allocated to each cable for their desired tension and directions through the Moore–Penrose inverse of the allocation matrix, which is determined by the connection points on the load~\cite{lee2017geometric}.
Some works further exploit the system redundancy in the null-space of the allocation matrix, offering capabilities for secondary tasks such as equal force distribution~\cite{geng2022load} and obstacle avoidance~\cite{wahba2024efficient, li2023nonlinear} while retaining the collective wrench on the load.
Once the required tension and cable directions are determined, a mid-level controller calculates the command thrust and attitude for each quadrotor, which are then executed by an inner-loop attitude controller.
However, despite considering the dynamic coupling effects, force-based methods are still far from fully exploiting the high agility of the cable-suspended system.
In fact, only low-speed (under ${1.5}~\mathrm{m/s}$) and low acceleration (under ${0.5}~\mathrm{m/s^2}$) flights have been successfully demonstrated in real-world experiments for load pose control using force-based approaches~\cite{li2023nonlinear,li2023rotortm,wahba2024efficient}.
Given that a loaded quadrotor with a thrust-to-weight ratio of 1.5 can easily reach an acceleration of over ${4}~\mathrm{m/s^2}$, existing solutions still largely compromise the inherent agility of quadrotors, making the cable-suspended multi-lifting system far from being able to operate in time-critical missions.

Here, we identify three major challenges obstructing the existing methods to achieve high agility in reality. 
First, the aforementioned force-based methods typically employ a cascaded control structure, which assumes that the load dynamics are substantially slower than those of the quadrotor.
This assumption fails during agile flights, where the load needs to change its pose rapidly.
With a cascaded control structure, the outer-loop commands can easily exceed the bandwidth of the inner loops, leading to instability, particularly in the presence of communication delays and actuator dynamics, which are often overlooked in simulation studies.

The second challenge is the high reliance on an accurate dynamic model, which is difficult to obtain.
The mismatch of the model, especially the mass and inertia of the payload, leads to an error in the thrust command sent to each quadrotor, ultimately causing tracking error and even instability.

The third challenge is the reliance on high-frequency load and cable measurements for closed-loop control, requiring 
additional sensors to be installed onto the load, such as reflective markers for a motion capture system ~\cite{li2023rotortm, wahba2024efficient, geng2022load}; or installing additional sensors on the quadrotors, such as downward-facing cameras~\cite{li2021cooperative}, cable tension sensors, and cable direction sensors~\cite{bernard2010load,bernard2011autonomous}.
These methods inherently suffer from sensor noise and latency and typically require nontrivial engineering efforts for installation and calibration, making them largely impractical for day-to-day real-world operations.

\subsection*{Trajectory-based framework}
In this article, we propose a trajectory-based framework to address the above challenges.
Our framework separates the controller into two submodules: an online kinodynamic motion planner and onboard trajectory tracking controllers. 
The kinodynamic motion planner considers the whole-body dynamics of the cable-suspended multi-lifting system, including the force-coupling effects, to generate dynamically feasible trajectories to each quadrotor in a receding-horizon fashion.
Then, a trajectory tracking controller is deployed onboard each quadrotor to generate the rotor-speed-level commands to follow the online-generated trajectories while considering the effect of the cable forces.

Specifically, we formulate the kinodynamic motion planner into a finite-time optimal control problem (OCP) which can be effectively solved within tens of milliseconds to generate predicted trajectories with a horizon of ${2}~\mathrm{s}$.
The OCP formulates safety-related constraints as path constraints, including thrust limitations, cable tautness, inter-quadrotor collision avoidance, and obstacle avoidance.
As the planner also takes into account the bandwidth and actuation constraints of the inner loop, the assumption of the time-scale separation principle required by existing solutions can be circumvented.
The generated trajectories include the full state of quadrotors along the horizon, hence our method allows the planner to run at a considerably lower frequency ($\leq$10Hz) than the outer-loop controllers of existing works ($\geq$100Hz).
This makes our method substantially more robust against the delay and noise on the state estimate of the load.

We deploy an estimator based on an extended Kalman filter (EKF) leveraging the load-cable dynamic model, quadrotor position and velocity estimates (generally available from an onboard state estimator), and accelerometers on quadrotors to provide satisfactory state estimates of the load and cables for the planner, achieving high-accuracy closed-loop tracking that outperforms state-of-the-art methods.
The onboard trajectory tracking controller employs the incremental nonlinear dynamic inversion (INDI) technique~\cite{smeur2016adaptive, tal2020accurate, sun2022comparative} and leverages the differential-flatness property of quadrotors~\cite{mellinger2011minimum} to follow the reference trajectories and instantly compensates for the forces from cables using measurements from the inertial measurement unit (IMU).
The mismatch in the planned cable tension that stems from the possible mismatch of the load inertia model is thereby effectively compensated for by the trajectory tracking controller, which eventually ensures high robustness against model uncertainties.

In the remainder of this article, we study the performance of the proposed trajectory-based framework in real-world experiments. The results reveal that the cable-suspended multi-lifting system controlled by our framework can achieve superior agility in pose control and trajectory following at high speeds (over ${5}~\mathrm{m/s}$) and accelerations (over ${8}~\mathrm{m/s^2}$). It can even rapidly change configurations to avoid obstacles and fly through narrow passages dynamically (Fig.~\ref{fig:eye_catcher}).
Our method also shows robustness against load model uncertainties, external wind disturbances, and quadrotor state estimation errors.
Moreover, the experiments were conducted without adding any sensors to the load to measure its pose, enhancing practicality in day-to-day real-world operations. 
The results and methods are summarized in Movie 1.

\section*{Results}
\subsection*{Experimental Setup}
We tested our algorithm through real-world experiments using three quadrotors to manipulate a ${1.4}~\mathrm{kg}$ payload. 
Each quadrotor, weighing ${0.6}~\mathrm{kg}$, experienced a substantial additional force due to the payload. 
Without loss of generality, we set the cable length to ${1}~\mathrm{m}$ for all quadrotors. 
These cables were attached to three distinct points on the rigid-body payload to enable pose control, with the other end connected to each quadrotor ${0.03}~\mathrm{m}$ below its center of gravity (CoG). 
The quadrotors were modified from the Agilicious open-source hardware platform ~\cite{foehn2022agilicious}, and each operates its onboard algorithms using a Raspberry Pi 5 mini PC. 
The centralized planner for our algorithm ran on a laptop at ${10}~\mathrm{Hz}$, sending commands to each quadrotor via WiFi.

We used motion capture systems to measure the poses of quadrotors at 100~Hz. 
These measurements were fused with onboard IMUs through an EKF to obtain state estimates of each quadrotor.
On the other hand, the state of the load for closed-loop control was estimated from the quadrotor states, agnostic to the sensors and state estimation algorithms onboard each quadrotor.
This also offers high practicality since no sensors are required to be attached to the load.
The effect of quadrotor state estimation error, typically seen in field operations without a motion capture system, is analyzed in the Section \textbf{Robustness against quadrotor state estimation error}.
A snapshot of the experimental setup is provided in Fig.~\ref{fig:experimental_setup}.

\subsection*{Agile Pose Control}

\begin{table}[t!]
\small
\centering
\caption{\textbf{Position tracking result.} Position root-mean-square-error (RMSE) in tracking references with different levels of agility. All reference trajectories were with a figure-eight shape. Our method substantially outperformed the two baseline methods (Geometric~\cite{lee2017geometric}, and NMPC~\cite{li2023nonlinear}), especially in tracking agile trajectories. The baseline methods were tested in a simulation environment, whereas our method was tested in both simulation and real-world experiments.}
\begin{tabular}{@{}lccc|cccc@{}}
\hline
Name of Ref. & $\textrm{vel}_{\mathrm{max}}$ & $\textrm{acc}_{\mathrm{max}}$ & $\textrm{jerk}_{\mathrm{max}}$ & Geometric \cite{lee2017geometric}   & NMPC \cite{li2023nonlinear}   & Ours  & Ours (\textbf{real-world})\\
trajectories & [m/s] & [m/s$^{2}$]& [m/s$^{3}$]& [m]& [m]& [m]& [m]\\ \hline
Slow       & 1      & 0.5    & 0.25   & 0.032         & 0.036    & \textbf{0.031}   & 0.102\\
Medium       & 2      & 2      & 2      & 0.135         & 0.159    & \textbf{0.067}   & 0.093          \\
Medium Plus       & 2      & 4      & 8      & Crash       & Crash  & \textbf{0.062}   & 0.117          \\
Fast       & 5      & 8      & 16     & Crash        & Crash  & \textbf{0.152}   & 0.197
\\ \hline
\end{tabular}
\label{tab: trajectry_tracking_comparison}
\end{table}

\begin{figure}
    \centering
    \includegraphics[width=0.95\linewidth]{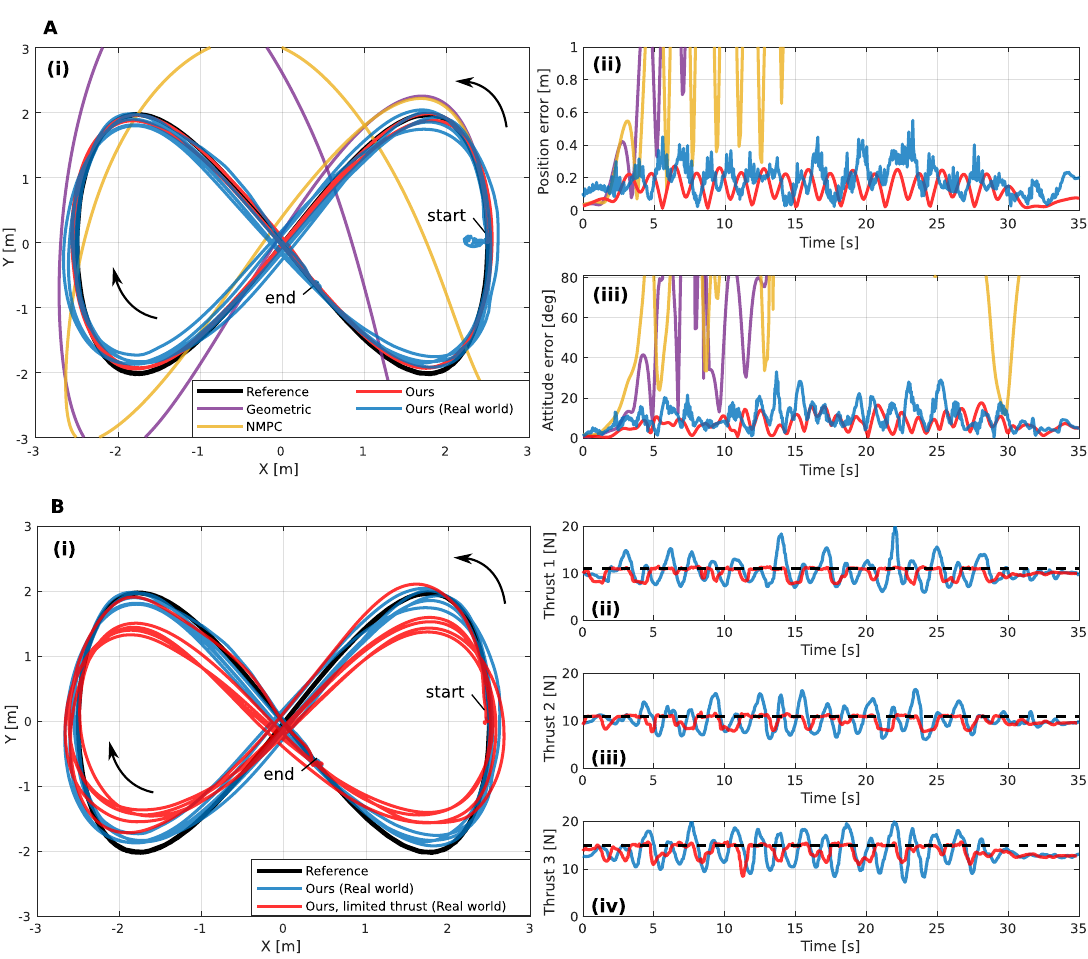}
    \caption{\textbf{Performance in tracking the reference Fast}. (\textbf{A}) Experiment comparing our method against two baseline methods to follow the reference trajectory Fast, a figure-eight trajectory with a maximum speed of 5~m/s and a maximum acceleration of 8~m/s$^2$. The detailed expression of the reference is given in Table~\ref{tab: algebratic_load_reference}. (i) Top view of the flight path of the CoG of the load.
    (ii-iii) Time history of the root-mean-square error of the load position and attitude tracking error of the load. We used axis-angle representation for the attitude error. (\textbf{B}) Experiment comparing our methods with tightened thrust limits and without, while tracking the reference Fast. (i) Top view of the flight path of the CoG of the load. Once the maximum thrust was limited, the reference trajectory became dynamically infeasible for the system to follow accurately (red). 
    (ii-iv) The commanded collective thrust of the three quadrotors with the reduced thrust limits (black dashed lines).}
    \label{fig:traj_tracking}
\end{figure}

To demonstrate that our method could control the cable-suspended multi-lifting system to achieve high agility, we tested its performance in tracking figure-eight trajectories with various levels of agility (increasing velocities, accelerations, and jerks), listed in Table~\ref{tab: trajectry_tracking_comparison}.
The algebraic expressions of the reference trajectories are given in Table~\ref{tab: algebratic_load_reference}.
At the same time, the reference heading also varied over time with a constant yawing rate of ${0.25}~\mathrm{rad/s}$. 
We present the results of our method obtained from real-world experiments.
We also present the results of our method compared with two baseline methods obtained in a simulation environment.
It should be noted that the parameters of the quadrotors and loads in these experiments were kept consistent between different approaches to ensure a fair comparison.

We selected two representative state-of-the-art methods as the baseline: geometric control~\cite{lee2017geometric, li2023rotortm} and NMPC~\cite{li2023nonlinear} that have been successfully demonstrated in real-world experiments.
These two force-based approaches both employ a conventional cascaded structure, namely using an outer-loop controller to generate the desired collective load wrench through a geometric controller or NMPC and distribute it to each quadrotor through an inner-loop controller.
Table~\ref{tab: trajectry_tracking_comparison} lists the tracking error in these reference trajectories.
Both baseline approaches could follow trajectories with relatively low agility (up until $v_\text{max}={2}~\mathrm{m/s}$, $a_\text{max}={2}~\mathrm{m/s^2}$).
However, they started to fail in following the reference Medium Plus, which involves higher peaks in acceleration and jerk, requiring rapid changes in the cables' directions to produce a fast time-varying wrench on the load.
Our method, by contrast, avoids using the cascaded structure employed by the baseline methods and thus can consequently allow fast variation of load pose and cable directions.
Therefore, it still successfully followed the reference Medium Plus and even the reference Fast, which has substantially larger accelerations and jerks.
A video recording of the comparison in simulation environments is provided in Movie S1.

Fig.~\ref{fig:traj_tracking}A presents the path and pose error while tracking the trajectory Fast that has a maximum velocity of ${5}~\mathrm{m/s}$ and a maximum acceleration of ${8}~\mathrm{m/s^2}$.  
The reference velocity and acceleration started from zero and gradually reached their maximum values.
As the reference velocity increased, both baseline methods failed to track the reference.
By contrast, our method succeeded in tracking the reference trajectory with a position-tracking root-mean-square-error (RMSE) of ${0.197}~\mathrm{m}$, and an attitude-tracking RMSE of ${12.9}~\mathrm{deg}$, in real-world experiments. 
The high closed-loop tracking accuracy came from the combined efforts of our controller and estimator. 
The time history of pose reference, estimate, and ground truth is presented in Fig.~\ref{fig:pose_estimation}.

Our algorithm considers dynamic coupling and thrust limits to prevent overloading the quadrotors.
In another experiment, we limited the maximum thrust of two quadrotors on the same side of the load from ${20}~\mathrm{N}$ to ${11}~\mathrm{N}$, and that of the third quadrotor on the opposite side to ${15}~\mathrm{N}$, since it needed to carry more lift.
Then we let the system track the reference Fast.
Consequently, the reference trajectory became dynamically infeasible for the thrust-limited multi-lifting system to follow precisely, leading to larger tracking error (0.363~m on position; 18.9~deg on attitude).
The tracking result is presented in Fig.~\ref{fig:traj_tracking}B.
Despite these thrust limitations, our method still enabled the multi-lifting system to follow the reference trajectory and avoid instability. 
Interestingly, our controller modulated trajectory curvature around turns to lower the required acceleration.
Throughout this process, our method ensured that the commanded thrust of each quadrotor was maintained within the reduced thrust limits.
Additionally, the variation in the collective thrust of each quadrotor was notably reduced with a tightened thrust limit, demonstrating that our method automatically adjusted the level of agility to match the capabilities of the quadrotors.

\subsection*{Obstacle Avoidance}

\begin{figure}
    \centering
    \includegraphics[width=1.0\linewidth]{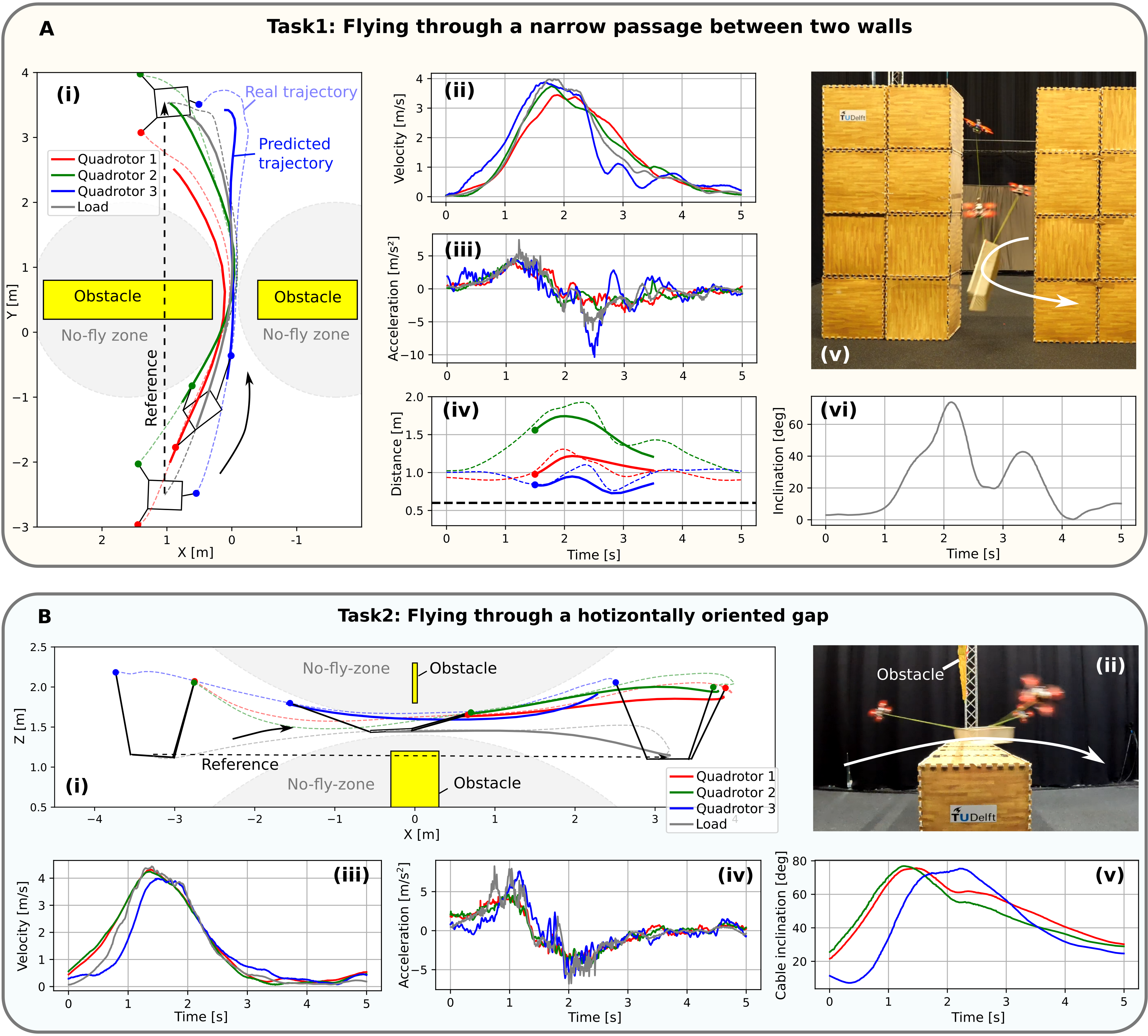}
    \caption{\textbf{Obstacle avoidance through dynamic motion.} 
    Both tasks were provided with a line segment reference that originally intersected the obstacles.  
    (\textbf{A}) Task 1: Flight through a narrow passage between two walls. (i) Top view of the load center and three quadrotors with predicted trajectories at $t={1.5}~\mathrm{s}$. (ii-iii) Velocity and acceleration profiles. (iv) Distances between quadrotors.  
    (v) Snapshot of the experiment when the multi-lifting system flew through the narrow passage. (vi) Load inclination during traversal, defined as the angle between the load-fixed z-axis and the world-frame z-axis. 
    (\textbf{B}) Task 2: Flight through a horizontally oriented narrow gap.  
    (i) Side view of the trajectory and predicted trajectories at $t={1.5}~\mathrm{s}$.  
    (ii) Snapshot of the experiment when the multi-lifting system flew through the horizontally oriented gap. (iii-iv) Velocity and acceleration profiles.  
    (v) Cable inclinations during traversal, defined as the angle between cable directions and the gravity.
    }
    \label{fig:obstacle}
\end{figure}

Our algorithm enabled high-speed obstacle avoidance without designing obstacle-free trajectories in advance, which required hundreds of seconds to generate for a multi-lifting system composed of more than three quadrotors~\cite{wahba2024efficient}.
Instead, we surrounded the obstacles with predefined no-fly zones and formulated them as second-order inequality constraints in the OCP. 
This way, the online-generated reference trajectories for the quadrotors ensured that the quadrotors and the load avoided the no-fly zones, thereby preventing collisions with the obstacles.

We demonstrated the obstacle avoidance capability of our algorithm in two previously unexplored challenging tasks.
In both tasks, the multi-lifting system had to navigate through gaps smaller than its original configuration size by leveraging its kinematic redundancy to reconfigure and squeeze through the narrow passage.
In the second task, traversal was performed dynamically, exploiting the momentum gained at high speed, since the system could not counteract gravity with its configuration at the moment of traversal if it attempted to fly statically.
A video of the obstacle avoidance experiments is provided in Movie S2.

\subsubsection*{Flying through a narrow passage}

In the first scenario, the multi-lifting system was commanded to fly through a narrow passage with a width of ${0.8}~\mathrm{m}$.
The width of the entire system in hovering condition was approximately ${1.4}~\mathrm{m}$, which was greater than the size of the gap.
We first commanded the quadrotors to carry the load to hover at an initial position.
Then we set a target at ${6}~\mathrm{m}$ away from the initial position along the y-axis of the inertial frame.
A minimum-snap reference trajectory~\cite{mellinger2011minimum} of the load was generated starting from the initial position to the target.
However, this reference trajectory intersected with the obstacle.
Without an obstacle avoidance mechanism, the system would have flown directly toward the wall and crashed.
Note that we used a constant orientation reference, with the load frame aligned with the world frame.

To guide the system through the opening, we defined two vertical cylinders as no-fly zones, each with a radius of ${1.5}~\mathrm{m}$, encompassing the real obstacles.
These two no-fly zones created a gap of ${0.2}~\mathrm{m}$ for the system to pass through, ensuring a ${0.3}~\mathrm{m}$ clearance from the real obstacles.
We selected several reference points on the load and on each quadrotor.
The algorithm then ensured that none of these reference points entered the no-fly zones.
Specifically, the reference points in this experiment were the center of each quadrotor and the four edges of the payload.
At the same time, the error between the actual and the reference pose was minimized in the cost function, encouraging the system to continue moving toward the final target pose.

Fig.~\ref{fig:obstacle}A presents the experimental data, illustrating the maneuvering process.
Our proposed algorithm generated predicted trajectories for both the load and the quadrotors at ${10}~\mathrm{Hz}$, allowing them to fly through the gap while adhering to the system kinodynamic model.
The planner automatically exploited the redundancy of the system to change the cable directions.
Since the width of the load (${0.54}~\mathrm{m}$) was greater than the gap between the two no-fly zones (${0.2}~\mathrm{m}$), the quadrotors managed to steer the load at a steep inclination of approximately ${70}~\mathrm{deg}$ to squeeze through the gap.
The distances between quadrotors were also included as constraints in the optimization problem.
Hence, their distances were kept greater than a safe margin (${0.8}~\mathrm{m}$) throughout the traversal.

Despite successfully avoiding the obstacles, the speed of the maneuver was not compromised.
The system reached a top speed of ${4}~\mathrm{m/s}$ during the fly-through maneuver, with a peak acceleration of over ${5}~\mathrm{m/s^2}$.
The load passed through the gap within ${1.2}~\mathrm{s}$ from the start of the maneuver and eventually stabilized at the target pose after successfully completing the traversal.

\subsubsection*{Flying through a horizontally oriented narrow gap}

In the second task, the quadrotors were commanded to carry the load through a horizontally oriented gap with a height of ${0.6}~\mathrm{m}$, while the height of the multi-lifting system in hover was around ${1.2}~\mathrm{m}$.
The experimental data of this fly-through maneuver is presented in Fig.~\ref{fig:obstacle}B.
In this case, we defined two horizontal cylinders as no-fly zones, each with a radius of ${4}~\mathrm{m}$, ensuring that the obstacles were encompassed within the no-fly zones with a minimum safety margin of ${0.2}~\mathrm{m}$.
The load was initially controlled to hover.
Next, a target pose behind the gap was sent to the algorithm, which generated a minimum-snap reference trajectory.
This trajectory, however, intersected with one of the obstacles.

Since the vertical size of the gap between the two no-fly zones was only ${0.2}~\mathrm{m}$, which was much smaller than the system’s hovering height of approximately ${1.2}~\mathrm{m}$ when all cables were nearly vertical, our algorithm controlled the quadrotors to spread out and stretch the cables, reducing the overall height of the system to enable it to pass through the gap.
During this process, the cable directions changed rapidly from nearly vertical to almost horizontal within ${1.2}~\mathrm{s}$.
At the moment of traversal, when the cables were nearly horizontal, the vertical components of the cable tensions could not compensate for the gravity of the load.
Therefore, the algorithm induced dynamic motions and took advantage of the momentum of the system to complete the fly-through.
In this maneuver, the load reached a maximum velocity of ${4}~\mathrm{m/s}$ and a peak acceleration of over ${7}~\mathrm{m/s^2}$, generating the momentum necessary for a successful traversal.
\subsection*{Robustness}
In this section, we investigate our method's robustness against load model uncertainties, external wind disturbances, and quadrotor state estimation errors.
These uncertainties are commonly seen in field operations.

\subsubsection*{Robustness against load model uncertainties}
\begin{figure}
    \centering
    \includegraphics[scale=0.9]{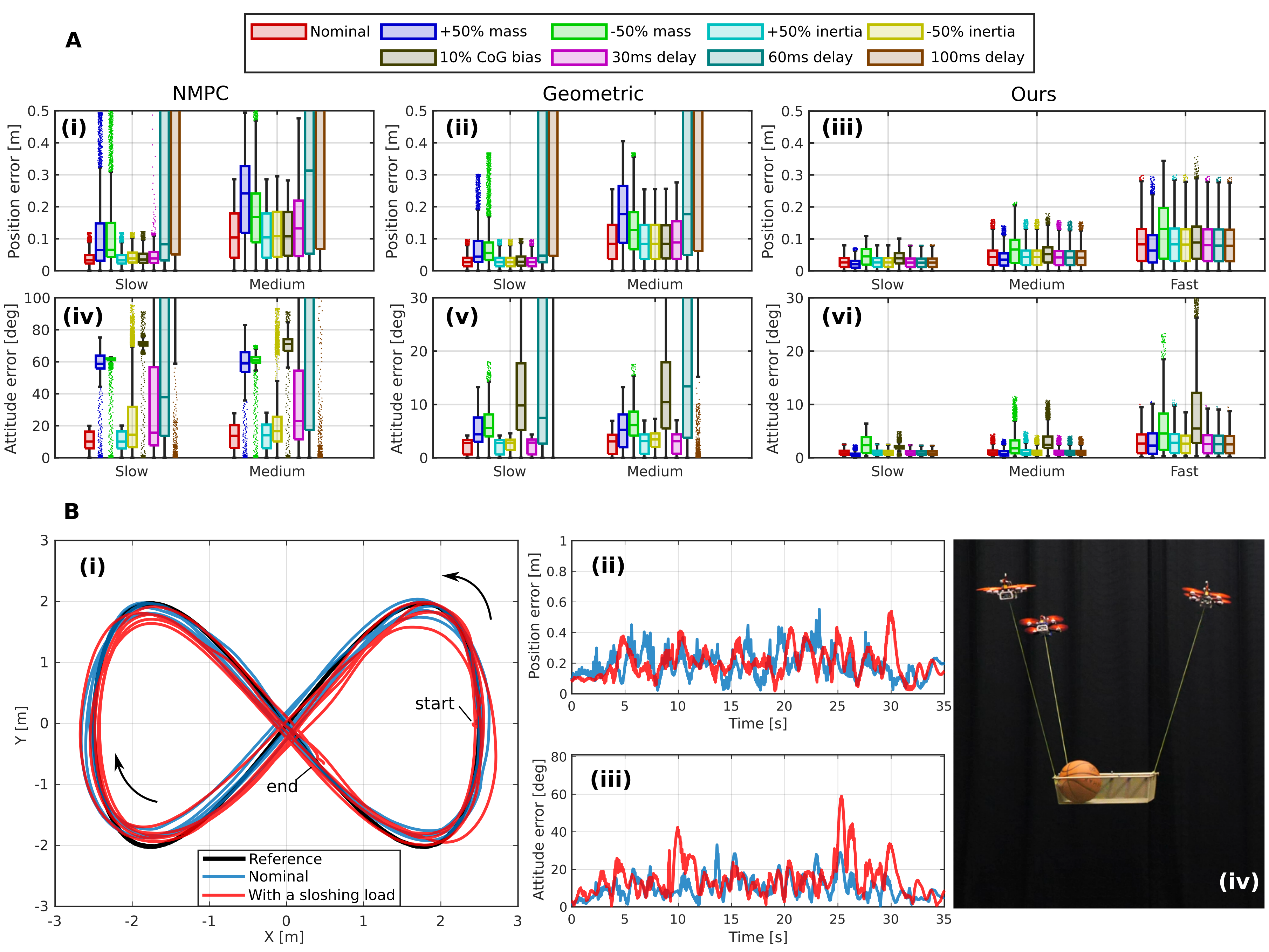}
    \caption{\textbf{Test under load model uncertainties and communication delays.} (\textbf{A}) Tracking performance of our method versus two baseline methods under various load model mismatches and communication delays while tracking reference Slow, Medium, and Fast as defined in Table~\ref{tab: algebratic_load_reference}. The baseline methods failed to follow the reference Fast even without mismatches, whereas our method remained robust. Each box corresponds to one run and summarizes the error at 4500 reference points: median (center line), 25th–75th percentiles (box), and whiskers extending to the minimum and maximum non-outlier values; outliers are defined as points lying beyond 1.5 interquartile range (IQR) from the box edges. (i–iii) Position error is in meters. (iv–vi) Attitude error is in degrees, calculated through axis-angle representations.
    (\textbf{B}) Real-world experiment where a ${0.6}~\mathrm{kg}$ basketball was placed onto the ${1.4}~\mathrm{kg}$ basket-shaped load and introduced a considerable inertia model mismatch of the load. Our method ran without knowing the presence of the basketball. (i) Top view of the path of the load CoG with and without a sloshing load. (ii-iii) Time history of the position and attitude tracking error. (iv) A snapshot of the experiment.}
    \label{fig:robustness}
\end{figure}

Being robust to uncertainties in the load model is a desirable feature, as it is often impractical to obtain an accurate load model during real-world operations.  
In Fig.~\ref{fig:robustness}A, we compared our method with the two baseline controllers in the presence of various types of model uncertainties on the load in a well-controlled simulation environment that quantified the model mismatch.  
The results showed that the baseline controllers were more sensitive to the model mismatch, especially in attitude control.  
By contrast, our approach sustained over 50\% mass and inertial mismatch in all the tests.  
These types of load model mismatch, commonly seen in practice, did not degrade the tracking performance, including the fastest reference trajectory given in Table~\ref{tab: trajectry_tracking_comparison}.  
Only the case with a 10\% bias of CoG led to an increase in attitude tracking error of about $5~\mathrm{deg}$.  
We also conducted simulations under step inputs of pose commands, which led to the same conclusion.  
The results are shown in Fig.~\ref{fig:step_response_robustness}.  

We further conducted a challenging real-world experiment, where we placed a ${0.6}~\mathrm{kg}$ basketball into the original basket-shaped load to introduce sloshing motion (Fig.~\ref{fig:robustness}B).
This led to a mass mismatch of 43\%, given that the mass of the original load was $1.4~\mathrm{kg}$.  
The motion of the basketball during flight also caused a pronounced time-varying CoG and inertia of the load if the basketball and the original load were considered as a single unit. 
We did not modify any parameters in the algorithm; the presence of the basketball was entirely unknown to our method.
Despite that, we commanded the multi-lifting system to follow the trajectory Fast.  
Since the sloshing inevitably introduced additional swaying motions, particularly during dynamic maneuvers, the tracking error with the sloshing load was slightly larger (${0.225}~\mathrm{m}$ vs. ${0.197}~\mathrm{m}$ for position RMSE and ${18.9}~\mathrm{deg}$ vs. ${12.9}~\mathrm{deg}$ for attitude).  
Nevertheless, our algorithm managed to control the multi-lifting system to follow the reference Fast with an unknown sloshing load, which the baseline methods could not achieve even with a perfect model (see Table~\ref{tab: trajectry_tracking_comparison}).  

\subsubsection*{Robustness against wind disturbance}

\begin{figure}
    \centering
    \includegraphics[width=1.0\linewidth]{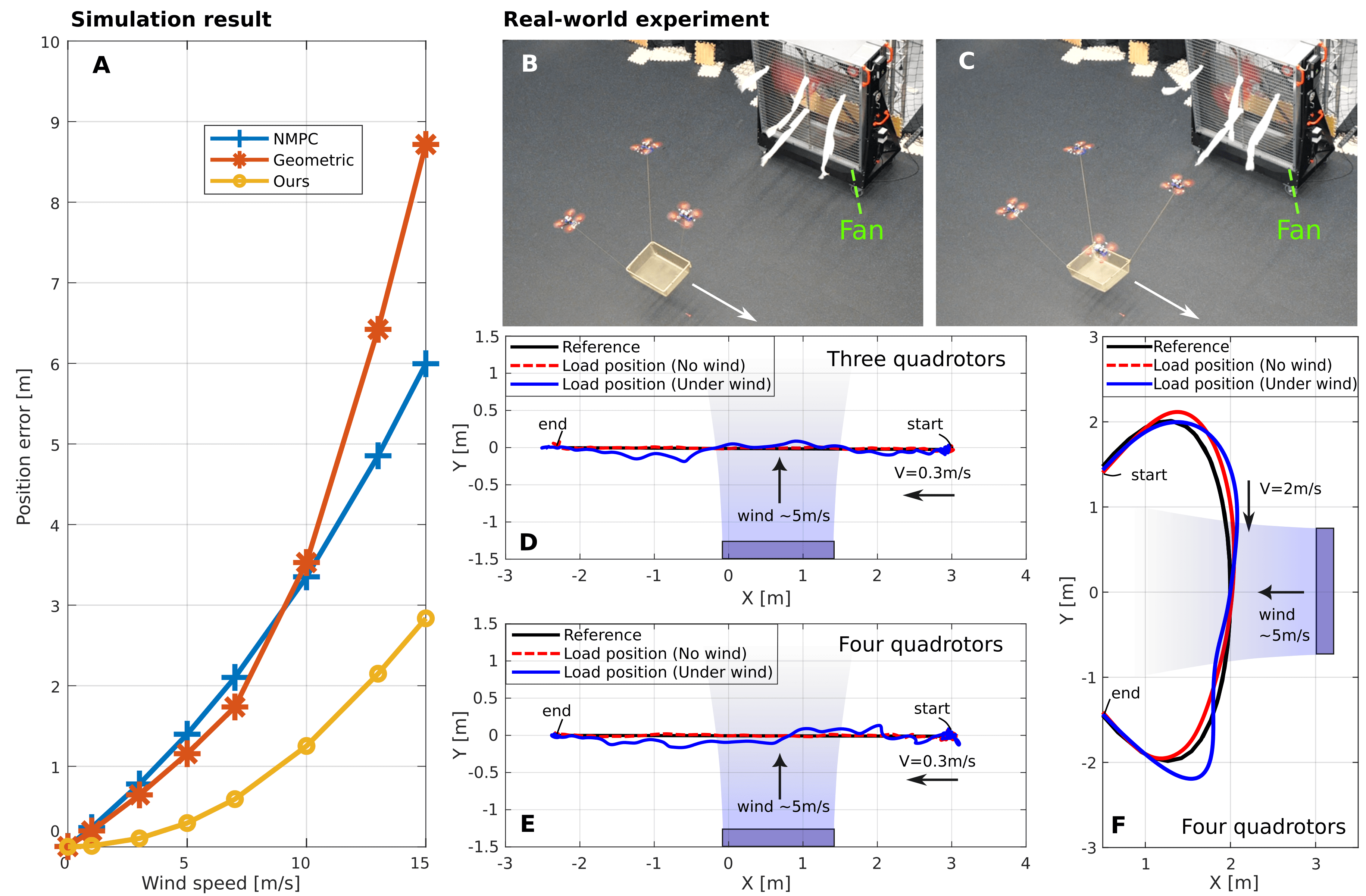}
    \caption{\textbf{Test under wind disturbances}. (\textbf{A}) Comparison of position error between our approach and the baseline methods at different wind speeds in simulation environments. (\textbf{B-C}) Snapshot of experiments with three and four quadrotors, respectively, under windy conditions generated by a 1.5 m diameter fan. (\textbf{D-E}) Real-world experimental data from three and four quadrotors carrying a load to follow a straight line at a speed of 0.3 m/s in a 5 m/s wind field. (\textbf{F}) Real-world experimental data with four quadrotors carrying the payload and flying over the wind field while following a curved trajectory at a speed of 2 m/s. The videos of the experiments are provided in Movie S4.}
    \label{fig:wind}
\end{figure}

We evaluated the performance of our method in both simulation and real-world experiments under windy conditions.  
In the simulation, we compared our method against the two baseline methods in the presence of various levels of wind.  
The system was commanded to hover at a target position.  
A horizontal wind was then ramped from zero to the designated speed over five seconds and persisted until the end of the simulation.  
We recorded the position error of the load once the system became stable.  
We employed the quadrotor drag model introduced in \cite{sun2022comparative} with parameters identified in real-world experiments.  
For the load, we employed a second-order drag model with a reference surface area of $0.05~\mathrm{m}^2$ and a drag coefficient of $1.05$ (value for a cube~\cite{hoerner1965}).  
Fig.~\ref{fig:wind}A presents the simulation results with wind up to $15~\mathrm{m/s}$ (greater than the maximum wind resistance of most commercial quadrotors), showing that our method notably outperformed the baselines.  
Note that the wind effect on the cables was neglected.  

We also conducted real-world experiments under a wind field of around $5~\mathrm{m/s}$ generated by a $1.5~\mathrm{m}$ diameter fan.  
We first commanded the multi-lifting system (consisting of three and four quadrotors, respectively) to follow a straight line across the wind field at only $0.3~\mathrm{m/s}$, exposing the system directly to the wind for a duration of 5 seconds.  
We then evaluated higher-speed flight at $2~\mathrm{m/s}$ by tracking a curved trajectory, which allowed us to assess performance under more dynamic conditions.  
In addition, we introduced the system into the wind field from an initially windless environment, creating a wind-gust-like scenario that further tested its ability to handle sudden changes in airflow.  

Fig.~\ref{fig:wind}D-F compares the top view of the trajectories under windy and windless conditions.  
Our framework enabled the multi-lifting system to operate at a moderate wind speed of $5~\mathrm{m/s}$.  
The disturbances acting on the quadrotors were effectively compensated for by the onboard flight controller.  
Since the aerodynamic model was not considered by the planner, the disturbance acting on the load led to greater tracking error compared to the case without wind disturbance ($0.048~\mathrm{m}$ vs. $0.055~\mathrm{m}$ for position RMSE with three quadrotors; $0.048~\mathrm{m}$ vs. $0.070~\mathrm{m}$ with four quadrotors).  
Such tracking errors could be further reduced in future work by identifying and integrating a wind-effect model into the centralized planner.  
Videos of the above experiments are provided in Movie~S4.  

\subsubsection*{Robustness against quadrotor state estimation error}
\label{sec:robustness_state_estimation}
\begin{figure}
    \centering
    \includegraphics[width=0.8\linewidth]{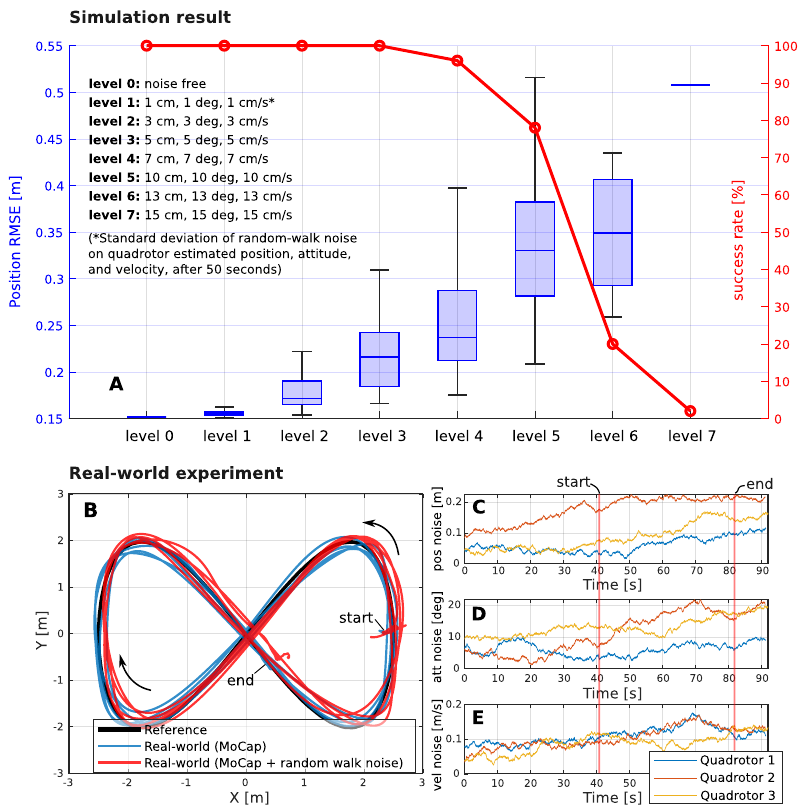}
    \caption{\textbf{Test under quadrotor state estimation errors.} (\textbf{A}) 
    Monte Carlo evaluation under varying levels of state estimation noise (50 runs per level), performed while tracking the Fast reference using noisy position, attitude, and velocity measurements. The red curve shows the success rate. The box plots summarize load position RMSE from the successful runs only: median (center line), 25th–75th percentiles (box), and whiskers extending to the minimum and maximum values.
    (\textbf{B}) Real-world flight test result under quadrotor state estimation error (level 3). We introduce random-walk noise on the original quadrotor state estimator that uses a motion capture system. (\textbf{C-E}) Time history of the quadrotor position, attitude, and velocity noise introduced in this flight test.}
    \label{fig:rand_walk_noise_state_estimation}
\end{figure}

Despite having no sensors on the load, our experiments still required state estimates from the quadrotors.  
Outside the lab environment without motion capture systems, greater estimation errors could occur due to imperfections in the quadrotor estimation algorithms and sensor noise from GPS, barometer, IMU, or onboard cameras.  
The quadrotor state estimation errors could deteriorate the closed-loop control performance of our framework.  
To test the sensitivity to quadrotor state estimation error, we deliberately added noise to the quadrotor position, attitude (Euler angle representation), and velocities.  
We selected random-walk noise on the quadrotor states, as it captured both the drift and stochastic fluctuations in the states.  
The level of random-walk noise was quantified by its standard deviation after 50 seconds starting from zero.  
We defined eight noise levels, each including 50 simulations where the system was commanded to follow the reference Fast.  

Fig.~\ref{fig:rand_walk_noise_state_estimation}A presents the Monte Carlo simulation results at each noise level, including the load position tracking RMSE and the success rate (i.e., no crash occurred).  
The closed-loop control performance gradually degraded as the noise level increased, which also reduced the success rate.  
Despite that, the success rate remained over 95\% even under noise level 4, where the standard deviations of position, attitude, and velocities were respectively $0.07~\mathrm{m}$, $7~\mathrm{deg}$, and $0.07~\mathrm{m/s}$ after 50 seconds of flight.  

A real-world experiment was also conducted by adding random-walk pose and velocity noise to the original state estimator output of each quadrotor.  
Fig.~\ref{fig:rand_walk_noise_state_estimation}B-E presents the estimation errors of all the quadrotors and the tracking performance compared to the case without added errors.  
The position errors of the quadrotors exceeded $0.1~\mathrm{m}$ over 90 seconds of flight, the attitude errors exceeded $10~\mathrm{deg}$, and the velocity errors exceeded $0.1~\mathrm{m/s}$, which was considerably less accurate than a commercial quadrotor operating in the field.  
Consequently, the position tracking error of the load increased from $0.197~\mathrm{m}$ to $0.278~\mathrm{m}$, and the attitude tracking error of the load increased from $12.9~\mathrm{deg}$ to $19.4~\mathrm{deg}$.  
Nevertheless, our method successfully controlled the multi-lifting system to follow the reference Fast without crashing.  
A video of the flights in simulation and in the real world under quadrotor state estimation error is provided in Movie~S5.  

\subsection*{Computational Load and Scalability}
\begin{figure}
    \centering
    \includegraphics[width=1.0\linewidth]{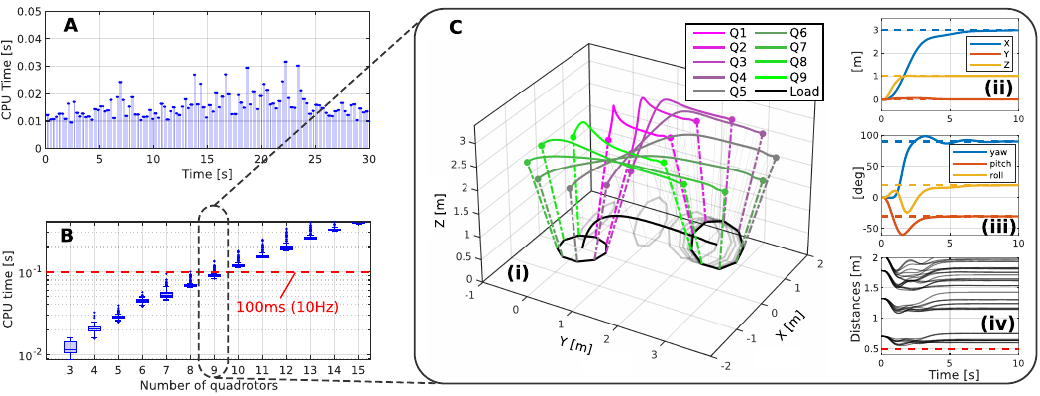}
    \caption{\textbf{Computational load and scalability.} (\textbf{A}) The CPU time to solve each OCP to follow the reference trajectory Fast with three quadrotors in real-world experiments. (\textbf{B}) Box plots of the CPU time of planning a trajectory with different numbers of quadrotors in a setpoint control task. Each box corresponds to data from 200 simulation steps: median (center line), 25th–75th percentiles (box), and whiskers extending to the minimum and maximum non-outlier values; outliers are defined as points lying beyond 1.5 IQR from the box edges. (\textbf{C}) Simulation result in a setpoint tracking task involving nine quadrotors. (i) 3D plot of the load and the CoG of the nine quadrotors. (ii-iii) Time history of the load pose (solid lines) compared with the reference pose (dashed lines). (iv) Time history of the distances between quadrotors (black solid lines) and the minimum distance allowed by our algorithm (red dashed line).}
    \label{fig:Scalability}
\end{figure}
Our method utilized a centralized structure, aiming to reach optimality in coordinating all the agents to manipulate the payload.  
The centralized structure, particularly the planner, posed a challenge to running in real time when multiple quadrotors were involved.  
The real-world experiments demonstrated that our method successfully coordinated three units in real time.  
Fig.~\ref{fig:Scalability}A presents the CPU time of the laptop running the planner (Intel Core i7-13700H) while tracking the trajectory Fast.  
The average CPU time consumed by the planner was ${15.3}~\mathrm{ms}$.  
Since the planner ran at ${10}~\mathrm{Hz}$ (i.e., every ${100}~\mathrm{ms}$), our algorithm only took 15.3\% of the computational budget.  

We further explored the potential of scaling up our algorithm to include more units.  
First, we successfully scaled up our method to four units in real-world experiments, as shown in Movie~S3.  
We also conducted simulations of setpoint tracking tasks with larger numbers of quadrotors.  
Without loss of generality, we assumed the load to be centrosymmetric, with cables connected to points evenly distributed around the load center at equal angular intervals depending on the number of quadrotors involved.  
The mass of the load was scaled proportionally with the number of quadrotors.  
We set the minimum distance constraint between quadrotors to be 1.6 times the distance between the corresponding contact points on the load, ensuring that constraint violations were possible if it was not imposed.  

As a result, Fig.~\ref{fig:Scalability}B shows that CPU time grew exponentially with the number of quadrotors included in the system.  
With our hardware setup, the planner supported nine units at 10~Hz without exceeding the compute budget.
Fig.~\ref{fig:Scalability}C presents the simulation results with nine quadrotors, including pose tracking, inter-quadrotor distances, and a 3D illustration.  
We believe that with a tailored optimizer for this particular problem, along with additional software and hardware optimizations, our method could be scaled up to substantially more units.  

\section*{Discussion}
Our experiments have shown that the proposed trajectory-based framework can substantially enhance the agility, robustness, and practicality of cable-suspended multi-lifting systems compared to the state-of-the-art.
To consider the dynamic coupling effects, the state-of-the-art framework requires a cascaded structure to coordinate and control multiple quadrotors to collaborate.
This conventional structure is built upon the principle of time-scale separation, assuming that quadrotors can instantly generate a resultant wrench on the load requested by an outer-loop controller, which limits the outer-loop gains to prevent instability and makes the tuning process tedious and task-specific~\cite{wahba2024efficient}.
Consequently, agility (high gain) and safety (low gain) are considered contradictory in the traditional framework.

In contrast, our method does not require the cascaded structure and addresses this issue by solving an online kinodynamic planning problem that considers the whole-body dynamics of the load-multi-quadrotor system.
The solution generates the states of the system over a future horizon, offering a predictive capability that allows for the inclusion of safety constraints during agile motion, rather than simply limiting gains in the traditional framework.
This key component of our solution enables precise pose control, trajectory tracking, and obstacle avoidance at high speeds (over ${5}~\mathrm{m/s}$) and accelerations (over ${8}~\mathrm{m/s^2}$).
Although the kinodynamic motion planning accounts for the whole-body dynamics, we have demonstrated that this problem is solvable on a midrange CPU in just a few milliseconds, enabling fast online generation to adjust to disturbances and avoid obstacles, and even with the potential to scale up to nine units with the current hardware and software setup.

We send the predicted trajectories to each quadrotor instead of a single reference point, offering two major advantages.
First, deploying a reference sampler together with a robust trajectory tracking controller on the quadrotors makes our framework substantially more robust to load model uncertainties, communication delays, and external wind disturbances.
This enables our algorithm to safely control the cable-suspended multi-lifting system, even when handling an unknown-mass sloshing load that brings over $40\%$ mismatch on the load mass model, or under moderate wind breeze at 5 m/s.
Second, sending a trajectory allows us to run the planner at over ten times lower frequency than traditional controllers, avoiding reliance on high-frequency measurements from sensors installed on the load.

We see several opportunities for future work.
Beyond the quadrotor multi-lifting problem, the trajectory-based framework has the potential to be applied to a wider range of robotic collaboration challenges, particularly those involving dynamic coupling, agility, and safety constraints.
Although our method guarantees high accuracy in the presence of load mismatch, underestimating the load mass and inertia can lead to violations of the maximum thrust constraints.
This problem can be alleviated by estimating the inertial properties online~\cite{petitti2020inertial}, or using the constrained tightening technique employed by a robust nonlinear optimal control framework, such as robust MPC ~\cite{richards2006robust, kohler2018novel}.
This requires a pre-estimation of the uncertainties and provides a more conservative, yet safer reference for quadrotors to avoid any violation of their dynamical constraints.
Another opportunity is combining our method with onboard perception algorithms in a GPS-denied environment.
This requires aligning coordinate frames across quadrotors, which is still a challenge for multi-agent perception algorithms, especially in agile flights. 
To support future works in this direction, we provide a preliminary analysis in the Supplementary Discussion to demonstrate the performance of our algorithm in the presence of misalignment between quadrotor coordinate frames.

Overall, our work paves the way for future aerial manipulation systems with substantially higher resilience, versatility, and agility to perform complex collaborative tasks in day-to-day operations, from search and rescue to precision delivery in difficult terrains.

\section*{Materials and Methods}
\begin{figure}
    \centering
    \includegraphics[width=1.0\linewidth]{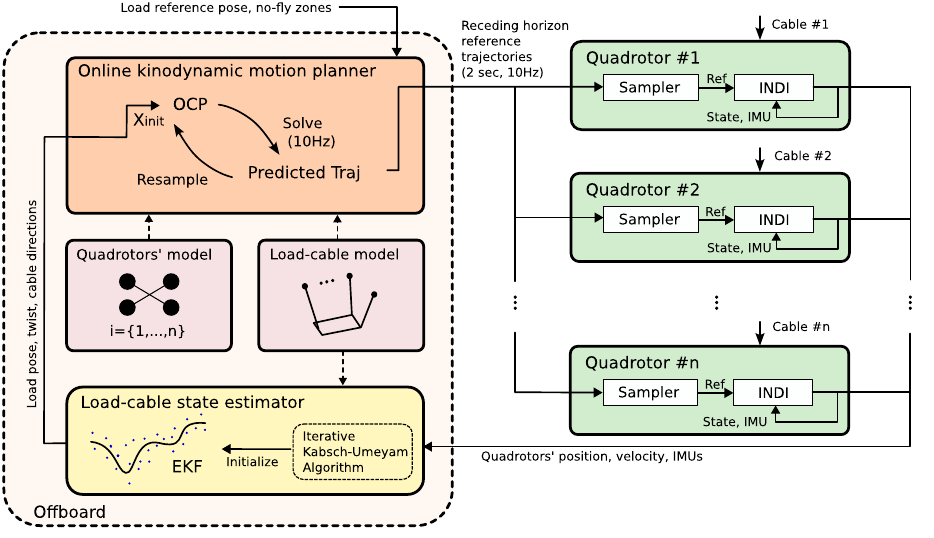}
    \caption{\textbf{Method overview.} Our framework includes a kinodynamic motion planner solving an OCP online at 10~Hz to generate receding-horizon reference trajectories of quadrotors given load reference pose and predefined no-fly zones. The OCP utilizes the whole-body dynamics of the system, including the quadrotor model and the load-cable model. The load's pose, twist, and cable directions are obtained from an EKF-based estimator. The remaining elements in the initial state of the OCP, namely the derivatives of the cable directions and tensions, are obtained by resampling the previously generated predicted trajectory to avoid oscillatory motion of the quadrotor when a new reference arrives. The load state estimator fuses the load-cable model, and the quadrotors' position, velocity, and IMU measurements to obtain estimates of load pose, twist, and cable directions. It is initialized through an iterative Kabsch-Umeyama algorithm given the initial quadrotor states. Onboard each quadrotor, a time-based sampler samples the received receding-horizon reference trajectory using the current timestamp to generate a single reference point, which is tracked by a trajectory tracking controller based on the INDI technique that regards the cable tensions as external disturbances, and compensates for them using the IMU measurements.}
    \label{fig:diagram}
\end{figure}

An overview of the method is shown in Fig.~\ref{fig:diagram}.
The proposed framework incorporates an optimization-based kinodynamic motion planner that generates real-time reference trajectories for the quadrotors.
It also includes a time-based sampler and an INDI-based trajectory tracking controller onboard each quadrotor.
In addition, the framework employs a centralized EKF to estimate the load pose and cable directions from the quadrotors’ position, velocity, and IMU measurements.
All modules are model-based and rely on the dynamic model of the cable-suspended multi-lifting system.
The following sections provide a detailed description of each module.
\subsection*{Modeling of cable-suspended multi-lifting systems}
\subsubsection*{Load-cable dynamic model}
The load-cable dynamic model describes the 6-DoF motion of the load and the motions of all the cables attached to the load.
Specifically, we employed the following definition of the state of the load-cable dynamic model:
\begin{equation}
    \boldsymbol{x}=[\boldsymbol{p},~\boldsymbol{v},~\boldsymbol{q},~\boldsymbol{\omega},~\boldsymbol{s}_1,~\boldsymbol{r}_1,~\dbm{r}_1,~\ddot{\boldsymbol{r}}_1,~t_1,~\dot{t}_1,...,~\boldsymbol{s}_n,~\boldsymbol{r}_n,~\dbm{r}_n,~\ddot{\boldsymbol{r}}_n
~t_n, \dot{t}_n]^\top,
\label{eq:load_cable_states}
\end{equation}
where $n$ is the number of quadrotors, $\boldsymbol{p}\in\mathbb{R}^3$, $\boldsymbol{v}\in\mathbb{R}^3$ are positions and velocities of the load, $\boldsymbol{q}\in \mathbb{S}^3$ is the unit quaternion describing the load attitude, $\boldsymbol{\omega}\in\mathbb{R}^3$ is the load angular velocity expressed in the load-fixed coordinate frame $\mathcal{F}_L$. 
Here the subscript $i$ indicates variables of the cable connected to the $i$-th quadrotor, where $\boldsymbol{s}_i\in \mathbb{S}^2$ is the cable direction pointing from the quadrotor to the load, $\boldsymbol{r}_i\in\mathbb{R}^3$ is the cable angular velocity, $t_i \in \mathbb{R}_{\geq 0}$ is the cable tension. 
An illustration of the reference frames and some symbols defined above can be found in Fig.~\ref{fig:symbol}.

We also adopted the following dynamic equations for the load:
\begin{equation}
\begin{split}
    \dbm{p} &= \boldsymbol{v},~~~~~
    \dbm{v} = -\sum_{i=1}^nt_i\boldsymbol{s}_i / m + \boldsymbol{g},\\
    \dbm{q} &= \frac{1}{2}\boldsymbol{\Lambda}(\boldsymbol{q}) 
\left[\begin{array}{c}
     0  \\
     \boldsymbol{\omega}
\end{array}\right],\\
    \boldsymbol{J}\dbm{\omega} &= -\boldsymbol{\omega}\times\boldsymbol{J}\boldsymbol{\omega}+\sum_{i=1}^nt_i\left(\boldsymbol{R}(\boldsymbol{q})^\top\boldsymbol{s}_i\times\boldsymbol{\rho}_i\right),
\end{split}
\label{eq: load_dynamics}
\end{equation}
where $\boldsymbol{J}\in\mathbb{R}^{3\times3}$ is the load inertia, $m$ is the load mass, $\boldsymbol{\rho}_i\in\mathbb{R}^3$ is the displacement of the $i$-th attachment point, expressed in the load frame, and $\boldsymbol{g}$ is the constant gravity vector. $\boldsymbol{\Lambda}(\boldsymbol{q})$ represents the quaternion multiplication, $\boldsymbol{R}(\boldsymbol{q})\in \mathrm{SO}(3)$ is the rotation matrix of the unit quaternion $\boldsymbol{q}$.
To ensure smooth quadrotor reference trajectories up to the jerk level, we employed the cable kinematic model, with the third-order derivative of cable angular velocity and the second-order derivative of cable thrust treated as bounded inputs, yielding
\begin{equation}
\begin{split}
        \dbm{s}_i &= \boldsymbol{r}_i\times\boldsymbol{s}_i,~~~ \dddot{\boldsymbol{r}}_i = \boldsymbol{\gamma}_i,~~~ \ddot{t}_i=\lambda_i,~~\textrm{for}~~~i=\{1,...,n\}, \label{eq:cable_kinematics}    
\end{split}
\end{equation}
where $\boldsymbol{\gamma}_i\in\mathbb{R}^3$ is the snap of cable directions, and $\lambda_i\in\mathbb{R}$ is the second-order derivative of cable tensions.
In the Supplementary Methods, we proved that the generated quadrotor trajectories are smooth up to the jerk level and also lead to a smooth angular velocity reference with states defined in the load-cable dynamic model in Equation~\ref{eq:load_cable_states}, as long as the $\boldsymbol{\gamma}_i$ and $\lambda_i$ are bounded.

\subsubsection*{Quadrotor dynamic model}
For the $i$-th quadrotor, we described the state space as $\boldsymbol{x}_i=[\boldsymbol{p}_i,~\boldsymbol{v}_i, ~\boldsymbol{q}_i, ~\boldsymbol{\omega}_i]^\top$, which corresponds to the CoG position $\boldsymbol{p}_i\in\mathbb{R}^3$, velocity $\boldsymbol{v}_i\in\mathbb{R}^3$, unit quaternion rotation $\boldsymbol{q}_i\in \mathbb{S}^3$, and angular velocity expressed in the quadrotor-fixed coordinate frame $\boldsymbol{\omega}_i\in\mathbb{R}^3$.
We employed the following rigid-body dynamics to derive the quadrotor equations of motion~\cite{foehn2021time}:
\begin{equation}
\begin{split}
            \dbm{p}_i &= \boldsymbol{v}_i,~~~~~
    \dbm{v}_i = (T_i\boldsymbol{z}_i + t_i\boldsymbol{s}_i + \boldsymbol{f}_{a,i})/m_i + \boldsymbol{g}, \\
    \dbm{q}_i &= \frac{1}{2}\boldsymbol{
    \Lambda
    }(\boldsymbol{q}_i)
\left[\begin{array}{c}
     0  \\
     \boldsymbol{\omega}_i
\end{array}\right],   \\
    \boldsymbol{J}_i\dbm{\omega}_i &=-\boldsymbol{\omega}_i\times\boldsymbol{J}_i\boldsymbol{\omega}+\boldsymbol{\tau}_i + \boldsymbol{\tau}_{a,i},
\end{split}
\label{eq:quadrotor_dynamics}
\end{equation}
where $m_i$ and $\boldsymbol{J}_i\in\mathbb{R}^{3\times3}$ are respectively the mass and the inertia matrix of the $i$-th quadrotor; 
$T_i \in \mathbb{R}_{\geq 0}$ is the collective thrust; $\boldsymbol{z}_i\in \mathbb{S}^2$ is the thrust direction aligning with the z-axis of the quadrotor body-fixed frame $\mathcal{F}_i$; $\boldsymbol{f}_{a,i}\in\mathbb{R}^3$ and $\boldsymbol{\tau}_{a,i}\in\mathbb{R}^3$ are the aerodynamic drag force and torque; $\boldsymbol{\tau}_i\in\mathbb{R}^3$ is the control torque generated by the rotors. 
\subsubsection*{Kinematic constraints}
We assumed that the cables’ tautness would be maintained by our algorithm throughout the operation, even during agile motions.
On the $i$-th quadrotor, the position of the cable contact point in the inertial frame is denoted as $\boldsymbol{p}_i$.
Then, the following kinematic constraint between the $i$-th quadrotor and load-cable dynamics holds:
\begin{equation}
    \boldsymbol{p}_i = \boldsymbol{p} + \boldsymbol{R}(\boldsymbol{q})\boldsymbol{\rho}_i - l_i\boldsymbol{s}_i ,
    \label{eq: kinematic constraint}
\end{equation}
where $l_i$ is the length of the cable.

\subsection*{Online kinodynamic motion planner}
\subsubsection*{Finite-time optimal control problem}
Our framework includes a centralized kinodynamic motion planner that generates smooth reference trajectories of all quadrotors in a receding-horizon fashion while considering the dynamic coupling between load and quadrotors.
Specifically, the planner is a discretized finite-time OCP, solved by a multi-shooting method~\cite{bock1984multiple}
\begin{equation}
\begin{split}
    \min J &= \sum_{k=0}^{N-1} \left(||\boldsymbol{x}_k-\boldsymbol{x}_{k,\text{ref}}||^2_{\boldsymbol{Q}} + |\boldsymbol{u}_k-\boldsymbol{u}_{k,\text{ref}}||^2_{\boldsymbol{R}}\right) \\
        &+ ||\boldsymbol{x}_N-\boldsymbol{x}_{N,\text{ref}}||^2_{\boldsymbol{P}} \\
    & \\
    \textrm{subject to}&~~~~~~~~~~\boldsymbol{x}_0 = \boldsymbol{x}_\text{init},~~~~\boldsymbol{x}_{k+1} = f\left(\boldsymbol{x}_k, \boldsymbol{u}_k\right),\\
    &~~~~~~~~~~\boldsymbol{h}(\boldsymbol{x}_{k+1}, \boldsymbol{u}_{k}) \leq \boldsymbol{0},~~~~k\in\{0,...,N\},
\end{split}
\label{eq:NLP}
\end{equation}
where the state equation uses the load-cable dynamic model (Equation~\ref{eq: load_dynamics} and \ref{eq:cable_kinematics}), and the input is $\boldsymbol{u} = \left[\boldsymbol{\gamma}_1,~\lambda_1,\ldots,\boldsymbol{\gamma}_n,~\lambda_n\right]$.
The quadrotor dynamics, albeit not explicitly included in the state equation for the reason of numerical efficiency, are included in the path constraints, particularly to avoid overloading quadrotors, and to perform obstacle avoidance.

The cost function is in a standard quadratic form to minimize the load pose tracking error and control effort for smoothness.
The reference states used in the cost function of the OCP are precomputed based on a polynomial load pose reference, with the remaining states derived using the flatness property of the cable-suspended multi-lifting system~\cite{sreenath2013dynamics}.
It is worth noting that in our experiments, the load reference did not consider avoiding obstacles.
Instead, we left the planner to decide and generate commands for the quadrotors to carry the load to avoid the obstacles.
In other words, the planner had the flexibility to deviate from the reference position or adjust the configurations to satisfy the obstacle avoidance constraints.
Naturally, our method could also follow load references generated by a higher-level offline planner (e.g.,~\cite{wahba2023kinodynamic}), other than the simple polynomial reference.

We discretized the horizon into $N=20$ non-equidistant segments, with intervals linearly increasing along the horizon.
Hence, it ensured higher fidelity of the predicted trajectory in the near future while extending the horizon length without increasing the number of discretization nodes.
This OCP was subsequently solved through the sequential quadratic programming (SQP) algorithm in a real-time iteration (RTI) scheme~\cite{diehl2002real}, implemented using ACADOS toolkit~\cite{verschueren2022acados}.
The solution of the OCP was the optimal input $\boldsymbol{u}_k^{*}$ and load-cable state $\boldsymbol{x}_k^{*}$ along the horizon
\begin{equation}
\begin{split}
    \boldsymbol{U^*} &= \left[\boldsymbol{u}^*_0,~\boldsymbol{u}^*_2,...,\boldsymbol{u}^*_{N-1}\right],\\
 \boldsymbol{X}^* &= \left[\boldsymbol{x}^*_1,~\boldsymbol{x}^*_2, ..., \boldsymbol{x}^*_N\right] = \boldsymbol{\pi}(\boldsymbol{U}^*,\boldsymbol{x}_\text{init}).
\end{split}
\end{equation}
Once the optimal state sequence $\boldsymbol{X}^*$ was obtained, we converted it to the position, velocity, acceleration, and jerk of the quadrotor through kinematic constraints (Equation~\ref{eq: kinematic constraint}) and its derivatives. 
It is worth noting that the headings of quadrotors, defined as the rotation angle around $\boldsymbol{z}_i$, did not affect the thrust directions or the motion of the load.
Therefore, we avoided explicitly setting the heading reference for the quadrotors but let the quadrotors maintain a zero yaw rate instead.

The OCP was then solved at a fixed frequency to generate a new trajectory online.
The initial state $\boldsymbol{x}_\text{init}$ in OCP was provided partially by the load-cable state estimator described in the Section \textbf{Load-cable state estimator}, and partially by resampling the trajectory from the latest OCP solution.
Specifically, we used the estimated load pose, twist, and cable directions to renew $\boldsymbol{x}_\text{init}$ to make sure the trajectories employ the up-to-date state of the load for closed-loop control.
The other states (cable rate, cable tensions, and their higher-order derivatives) were directly estimated by resampling on the previously generated trajectory.
We observed that this treatment has two benefits. First, it ensured smooth transitions between consecutive reference trajectories, avoiding any abrupt and jerky maneuvers by the quadrotors.
It also avoided numerical differentiation of state estimator values, which is usually impractical due to the high requirements for accuracy and smoothness in the estimator output.
\subsubsection*{Path constraints}
We included several path constraints $\boldsymbol{h}(\boldsymbol{x})\leq\boldsymbol{0}$ in the OCP to ensure safety. 
To avoid overloading each quadrotor, we included the thrust constraint:
\begin{equation}
    T_{i,\text{min}}\leq T_i(\boldsymbol{x}) \leq T_{i,\text{max}},
\end{equation}
where $T_i(\boldsymbol{x})$ is the thrust of each quadrotor as a function of the state of the load-cable dynamic model.
Specifically, $T_i(\boldsymbol{x})$ was obtained through quadrotor dynamics (Equation~\ref{eq:quadrotor_dynamics}),
\begin{equation}
    T_i(\boldsymbol{x}) = \big\|\left(\dbm{v}_i(\boldsymbol{x}) - \boldsymbol{g}\right)m_i-t_i\boldsymbol{s}_i - \boldsymbol{f}_{a,i}\big\|,
    \label{eq:T_i}
\end{equation}
where $\dbm{v}_i(\boldsymbol{x})$ was calculated from the second-order derivative of the kinematic constraints (Equation~\ref{eq: kinematic constraint}).

We assumed that the cables would remain taut throughout the operation. 
Therefore, a cable tension constraint is included.
\begin{equation}
    0 < t_\text{min}\leq t_i\leq t_\text{max},
\end{equation}

To avoid inter-quadrotor collisions, the minimum distance constraints were also provided for every pair of quadrotors indexed by $i$ and $j$
\begin{equation}
    0 < d_\text{min} \leq ||\boldsymbol{p}_i(\boldsymbol{x}) - \boldsymbol{p}_j(\boldsymbol{x})||,
\end{equation}
where $d_\text{min}$ is the predefined minimum distance, $\boldsymbol{p}_i(\boldsymbol{x})$ and $\boldsymbol{p}_j(\boldsymbol{x})$ are the positions of the $i$-th and the $j$-th quadrotor.
 
We also established several control points on the system to ensure it avoided obstacles.
Without loss of generality, we used the CoG of each quadrotor together with the attaching points on the load.
For each obstacle, and each control point denoted by $\boldsymbol{p}_c(\boldsymbol{x})$, the following constraint ensured that none of the control points entered the no-fly zone encompassing the obstacle
\begin{equation}
   d^2_{o, \mathrm{min}}\leq (\boldsymbol{p}_c(\boldsymbol{x}) - \boldsymbol{p}_o)^\top\boldsymbol{C}(\boldsymbol{p}_c(\boldsymbol{x}) - \boldsymbol{p}_o),
\end{equation}
where $\boldsymbol{C}\in\mathbb{R}^{3\times3}$ is a diagonal matrix controlling the shape of the no-fly zone, $\boldsymbol{p}_o$ is the center of the no-fly zone, $d_{o,\mathrm{min}}$ is the safe distance from the control points to the center.
The position and shape of the obstacle can be determined either offline or detected online.
In the case when the obstacle is detected online, we can set up the problem to support a large number of obstacles and set the inactive obstacle constraints with a zero radius or place them far from the current positions. 
Once a new obstacle is detected, we activate one of the reserved inequality constraints by adjusting its parameters to match the size and location of the detected obstacle.

Lastly, to ensure the bounded input to the OCP, the control input constraint $\boldsymbol{u}_\text{min}\leq\boldsymbol{u}\leq\boldsymbol{u}_\text{max}$ was imposed in the experiments.

It is worth noting that all path constraints were inequality constraints and handled using slack variables to ensure problem feasibility.

\subsection*{Load-cable state estimator}
\label{sec:estimator}
To update $\boldsymbol{x}_\text{init}$ in the OCP of the planner, the load pose, twist, and cable directions must be estimated in real-time.
Instead of relying on additional sensors such as downward-facing cameras~\cite{li2021cooperative} or adding motion capture markers on the load~\cite{li2023rotortm} in the state-of-the-art approaches, our proposed estimator utilizes only the quadrotors' states and the load-cable dynamics. This eliminates the need for any hardware modifications.

We chose EKF to solve this state estimation problem thanks to its simplicity and computational efficiency.
We omit the detailed steps in EKF and only describe the selections of states, measurements, models, and initialization.

The state vector of EKF comprised pose and twist of the load, as well as the positions and velocities of all quadrotors, namely $\hat{\boldsymbol{x}}=\left[\boldsymbol{p},~\boldsymbol{v},~\boldsymbol{q},~\boldsymbol{\omega},~\boldsymbol{p}_1,~\boldsymbol{v}_1,...,~\boldsymbol{p}_n,~\boldsymbol{v}_n\right]^\top$.
State prediction was performed using the load dynamics and quadrotor dynamics (Equations~\ref{eq: load_dynamics} and \ref{eq:quadrotor_dynamics}).
The cable directions to solve the load dynamic were obtained through the kinematic constraint (Equation~\ref{eq: kinematic constraint}).
The cable forces in these equations were estimated through a spring-damper model, i.e., $t_i = k_\text{stiff}d_i + k_\text{damp}\dot{d}_i$ where $d_i$ is the distance between the position of the $i$-th quadrotor and its connection point on the load, namely $d_i = ||\boldsymbol{p}_i - \boldsymbol{R}(\boldsymbol{q})\boldsymbol{\rho}_i-\boldsymbol{p}||$; $k_\text{stiff}$ and $ k_\text{damp}$ are positive coefficients.

The EKF took the cables' directions, together with the quadrotors' positions and velocities as measurements, namely $\tilde{\boldsymbol{y}} =\left[\tilde{\boldsymbol{s}}_1, \tilde{\boldsymbol{p}}_1, \tilde{\boldsymbol{v}}_1,...,\tilde{\boldsymbol{s}}_n, \tilde{\boldsymbol{p}}_n, \tilde{\boldsymbol{v}}_n\right]^\top$.
The quadrotor positions and velocities, and their covariances, were obtained directly from their onboard state estimators.
The cable directions were obtained indirectly from the accelerometer sensor that is commonly available on a quadrotor drone.
Since the accelerometer directly measures the specific force (the mass-normalized force excluding gravity), it captures the combined forces, including the cable tension, aerodynamic drag, wind force, and rotor thrusts.
For each quadrotor, we identified a collective thrust model $\bar{T}_i$, and a drag model $\bar{\boldsymbol{f}}_{a,i}$ with the following expressions:
\begin{equation}
    \bar{T}_i = \sum_{j=1}^4c_t\omega_{j,i}^2,~~~~~\bar{\boldsymbol{f}}_{a,i}=\boldsymbol{R}(\boldsymbol{q}_i)\boldsymbol{D}_a\boldsymbol{R}(\boldsymbol{q}_i)^\top\boldsymbol{v}_i,
\end{equation}
where $c_t$ is the thrust coefficient of the rotors, $\omega_{j,i}$ is the rotor speed, $\boldsymbol{D}_a\in\mathbb{R}^{3\times3}$ is the aerodynamic coefficient matrix~\cite{faessler2017differential}.
According to the quadrotor dynamics (Equation~\ref{eq:quadrotor_dynamics}), subtracting them from the accelerometer readings provides the force vector from cables.
Then the cable directions were approximated by
\begin{equation}
    \tilde{\boldsymbol{s}}_{i} = (m_i\boldsymbol{a}_i - \bar{T}_i\boldsymbol{z}_i - \bar{\boldsymbol{f}}_{a,i})/||m_i\boldsymbol{a}_i - \bar{T}_i\boldsymbol{z}_i - \bar{\boldsymbol{f}}_{a,i}||,
\end{equation}
where $\boldsymbol{a}_i=\dbm{v}-\boldsymbol{g}$ is the unbiased accelerometer measurement.  

We observed that the covariance matrix of the cable direction was challenging to determine directly from the accelerometer properties, as it depends on the accuracy of $\bar{T}_i$, $\bar{\boldsymbol{f}}_{a,i}$, and is also affected by wind disturbance.
Therefore, we tuned this covariance experimentally. 
We observed that underestimating the covariance could result in a noisy load pose estimate, whereas overestimating it led to an excessively free motion in the load estimate.

We initialized the EKF with a first-order guess of load pose and twist, and cable directions.
We assumed a static initial load state, namely a zero twist.
As for the load pose and cable directions, we propose an algorithm to provide a guess iteratively through the Kabsch-Umeyama algorithm~\cite{umeyama1991least}.
The details of the algorithm are provided in Algorithm~S1.

\subsection*{Trajectory tracking controller on quadrotors}
In our setup, every ${100}~\mathrm{ms}$, the most recently generated reference trajectories by the planner were sent to each quadrotor, and then followed by a differential-flatness-based trajectory tracking controller deployed onboard, modified from~\cite{sun2022comparative}.
Since the trajectory tracking controller operated at a higher frequency (${300}~\mathrm{Hz}$) than the interval between nodes in the reference trajectory, a time-based sampler was implemented to generate high-frequency reference states by linearly interpolating between the discretized nodes of the reference trajectory. 
The sampler continued to sample along the reference trajectory until a new reference was received.

The onboard trajectory tracking controller then computed the thrust command, including magnitude $T_{i,\mathrm{des}}$ and direction $\boldsymbol{z}_{i,\mathrm{des}}$, through the following PD controller:
\begin{equation}
\begin{split}
        T_{i,\text{des}}\boldsymbol{z}_{i,\text{des}}/m_i &= \boldsymbol{K}_p\left(\boldsymbol{p}_{i,\text{ref}}-\boldsymbol{p}_i\right) + \boldsymbol{K}_v\left(\boldsymbol{v}_{i,\text{ref}}-\boldsymbol{v}_i\right) + \dot{\boldsymbol{v}}_{i,\text{ref}} + 
        \boldsymbol{f}_\text{ext}/ m_i,
\end{split}
\end{equation}
where $\boldsymbol{K}_p\in\mathbb{R}^{3\times3}$ and $\boldsymbol{K}_v\in\mathbb{R}^{3\times3}$ are positive definite gain matrices; $\boldsymbol{f}_\textrm{ext}$ are external forces represents external forces on the quadrotor, excluding thrust and gravity, namely cable tension, aerodynamic drag, and wind.
We estimated external forces using the accelerometer on the quadrotor through the relationship $\boldsymbol{f}_\text{ext} = m_i \boldsymbol{a}_{i,\text{filtered}} - \boldsymbol{f}_{i,\text{filtered}}$, where $\boldsymbol{a}_{i,\text{filtered}}$ is the unbiased and low-pass-filtered accelerometer measurement, $\boldsymbol{f}_{i,\text{filtered}}$ is the current collective thrust denoised with the same filter.
Then, we used a tilt-prioritized attitude controller~\cite{brescianini2018tilt} to generate the angular acceleration command $\boldsymbol{\alpha}_{i,\mathrm{des}}$ from the desired attitude $\boldsymbol{z}_{i,\text{des}}$, the reference jerk, and the zero yaw rate reference. 

The angular acceleration and force commands were subsequently allocated to rotor speed commands through an INDI inner-loop controller, which is a sensor-based adaptive controller robust against external torque disturbances such as aerodynamic torque, motor differences, quadrotor CoG bias, etc.
Finally, the rotor speed commands generated by INDI were sent to the electronic speed controllers (ESCs) through the DShot protocol.
We refer interested readers to ~\cite{foehn2022agilicious} for further details about the hardware implementations.
We also provide key equations for INDI in the Supplementary Methods.


\clearpage 

%
\bibliography{references} 
\bibliographystyle{sciencemag}

%
%
%
%
%
%


\section*{Acknowledgments}
The authors would like to thank Kseniia Khomenko, Maurits Pfaff, Dr. Gianluca Corsini, Riccardo Belletti, Dr. Dennis Benders, and Shantnav Agarwal for their support during the experiments.
We also extend our gratitude to Dr. Yunlong Song, Dr. Guanrui Li, Prof. Guido de Croon, and Prof. Davide Scaramuzza for their valuable discussions and feedback.
\paragraph*{Funding:}
This work was supported by the Dutch Research Council (NWO) through the Veni Talent Programme (Grant No. 20256, Accurate Aerial Manipulation), by the French National Research Agency (ANR) under Grant ANR-24-CE33-5799 (MATES), and by the European Union’s Horizon 2020 Research and Innovation Programme under Grant Agreement No. 871479 (AERIAL-CORE).
\paragraph*{Author contributions:}
S.S. acquired funding, formulated the main ideas, developed the algorithm, conducted experiments, and wrote the manuscript.
X.W. developed the algorithm, conducted experiments, and revised the manuscript. 
D.S helped with the algorithm implementation and revised the manuscript.
A.F, M.T, J.A formulated ideas, acquired funding, and revised the manuscript.
\paragraph*{Competing interests:}
There are no competing interests to declare.
\paragraph*{Data and materials availability:}
All (other) data needed to evaluate the conclusions in the paper are present in the paper or the Supplementary Materials. The data for this study have been deposited in the database Dryad (https://doi.org/10.5061/dryad.n2z34tn6w).


\subsection*{Supplementary materials}
Supplementary Methods\\ 
Supplementary Discussion\\ 
Figures S1 to S5\\
Table S1\\
Algorithm S1\\
Movies S1 to S5\\


\newpage


\renewcommand{\thefigure}{S\arabic{figure}}
\renewcommand{\thetable}{S\arabic{table}}
\renewcommand{\theequation}{S\arabic{equation}}
\renewcommand{\thepage}{S\arabic{page}}

\setcounter{figure}{0}
\setcounter{table}{0}
\setcounter{equation}{0}
\setcounter{page}{1} 


\begin{center}
\section*{Supplementary Materials for\\ \scititle}


Sihao Sun$^{\ast}$,
	Xuerui Wang,
	Dario Sanalitro,\\
        Antonio Franchi,
        Marco Tognon,
        Javier Alonso-Mora\\ 
\small$^\ast$Corresponding author. Email: sihao.sun@outlook.com\\
\end{center}

\subsubsection*{This PDF file includes:}
Supplementary Methods \\
Supplementary Discussion \\
Figures S1 to S5\\
Table S1\\
Algorithm S1\\
Captions for Movies S1 to S5\\

\subsubsection*{Other Supplementary Materials for this manuscript:}
Movies S1 to S5\\

\newpage


\section*{Supplementary Methods}
\subsection*{Proof of Smoothness of Quadrotor Trajectories}
\label{sec:proof_smoothness}
\textbf{Proposition 1:} 
\textit{When the $i$-th cable remains taut, the trajectory of $i$-th quadrotor, denoted as $\boldsymbol{p}_i(t)$, is $C^3$ smooth if $\boldsymbol{\lambda}_i$ and $\boldsymbol{\gamma}_i$ defined in Equation~\ref{eq:cable_kinematics} are bounded.}

\noindent
\textbf{Proof:} 
For the $i$-th quadrotor, the kinematic constraint (Equation~\ref{eq: kinematic constraint}) holds when the corresponding cable is taut.
Then we take the 3rd-order derivative of Equation~\ref{eq: kinematic constraint} to obtain the jerk of the quadrotor

\begin{equation}
\begin{split}    
    \ddbm{v}_i  = &~\ddbm{v} + \boldsymbol{R}(\boldsymbol{q})\biggl\{\boldsymbol{\omega}\times\biggl[\dbm{\omega}\times\boldsymbol{\rho}_i+ \boldsymbol{\omega}\times(\boldsymbol{\omega}\times\boldsymbol{\rho}_i)\biggr]  \\
    & + \ddbm{\omega}\times\boldsymbol{\rho}_i+\dbm{\omega}\times(\boldsymbol{\omega}\times\boldsymbol{\rho}_i) + \boldsymbol{\omega}\times(\dbm{\omega}\times\boldsymbol{\rho}_i)\biggr\}  \\
    & - l_i\biggl\{\ddbm{r}_i\times\boldsymbol{s}_i + 2\dbm{r}_i\times(\boldsymbol{r}_i\times\boldsymbol{s}_i)+\boldsymbol{r}_i\times(\dbm{r}_i\times\boldsymbol{s}_i) \\
    &+ \boldsymbol{r}_i\times\biggl[\boldsymbol{r}_i\times(\boldsymbol{r}_i\times\boldsymbol{s}_i)\biggr]\biggr\},
\end{split}        
\end{equation}
where $\dbm{\omega}$ is given in the load dynamics (Equation~\ref{eq: load_dynamics}); $\ddbm{v}$ and $\ddbm{\omega}$ are obtained by taking the derivative of both sides of Equation~\ref{eq: load_dynamics}:
\begin{equation}
    \ddbm{v} = - \frac{1}{m}\sum_{i=1}^n\biggl[\dot{t}_i\boldsymbol{s}_i+t_i(\boldsymbol{r}_i\times\boldsymbol{s}_i)\biggr],
\end{equation}
\begin{equation}
    \ddbm{\omega} = \boldsymbol{J}^{-1}\biggl\{ -\dbm{\omega}\times\boldsymbol{J}\boldsymbol{\omega} - \boldsymbol{\omega}\times\boldsymbol{J}\dbm{\omega} + \sum_{i=1}^n\biggl[\dot{t}_i\boldsymbol{R}^\top\boldsymbol{s}_i+t_i\biggl(-\boldsymbol{\omega}\times\boldsymbol{R}^{\top}\boldsymbol{s}_i + \boldsymbol{R}^{\top}(\boldsymbol{r}_i\times\boldsymbol{s}_i)\biggr)\biggr]\times\boldsymbol{\rho}_i\biggr\},
\end{equation}
Judging from the load-cable dynamics (Equation~\ref{eq: load_dynamics} and ~\ref{eq:cable_kinematics}), the continuousness of $\ddbm{v}_i$ is determined by the highest-order states $\dot{t}_i$ and $\ddbm{r}_i$.
Therefore, $\ddbm{v}_i$ is continuous; namely $\boldsymbol{p}_i$ is $C^3$-smooth, when $\ddot{t}_i=\gamma_i$ and $\dddot{\boldsymbol{r}}_i=\boldsymbol{\lambda}_i$ are bounded.
\hfill $\square$

The OCP of the planner takes $\gamma_i$ and $\boldsymbol{\lambda}_i$ as inputs, which can be bounded by setting input constraints.
Hence, the generated trajectories of all quadrotors are smooth up jerk as long as the cable tautness is guaranteed.
One step further, once the reference jerk is continuous, we can also obtain a smooth angular velocity reference.

\noindent
\textbf{Proposition 2:} 
\textit{The angular velocity of the $i$-th quadrotor expressed in the inertial frame, denoted by $\boldsymbol{\omega}_i^{\mathcal{I}}\in\mathbb{R}^3$ is $C^0$-smooth if $\boldsymbol{\lambda}_i$ and $\gamma_i$ defined in Equation~\ref{eq:cable_kinematics} are bounded, and aerodynamic drag $\boldsymbol{f}_{a,i}$ is at least $C^1$-smooth.}

\noindent
\textbf{Proof:} 
To obtain the angular velocity reference of each quadrotor, we need to revisit the translational dynamic equation of the $i$-th quadrotor
\begin{equation}
    \dbm{v}_i = \left(T_i\boldsymbol{z}_i+t_i\boldsymbol{s}_i+\boldsymbol{f}_{a,i}\right)/m_i + \boldsymbol{g}
    \label{eq:quadrotor_trans_dyn}
\end{equation}
Taking the derivative of both sides of Equation~\ref{eq:quadrotor_trans_dyn}, we have
\begin{equation}
\begin{split}
    \boldsymbol{h}_i &\triangleq \boldsymbol{\omega}^\mathcal{I}_i\times\boldsymbol{z}_i \\
    & = \biggl[m_i\ddbm{v}_i-\dot{T}_i\boldsymbol{z}_i-\dot{t}_i\boldsymbol{s}_i-t_i\left(\boldsymbol{r}_i\times\boldsymbol{s}_i\right) -\dbm{f}_{a,i}\biggr]/T_i
\end{split}
\end{equation}
Since the yaw rate references are zero for all quadrotors, $\boldsymbol{\omega}^\mathcal{I}_i$ is perpendicular to $\boldsymbol{z}_i$.
Then we can obtain the expression of $\boldsymbol{\omega}^\mathcal{I}_i$ by
\begin{equation}
    \boldsymbol{\omega}^\mathcal{I}_i = \boldsymbol{z}_i\times\boldsymbol{h}_i = \frac{1}{T_i} \boldsymbol{z}_i\times\biggl[\ddbm{v}_i - \dot{t}_i\boldsymbol{s}_i-t_i(\boldsymbol{r}_i\times\boldsymbol{s}_i) -\dbm{f}_{a,i}\biggr]
    \label{eq:angular_rate_reference}
\end{equation}
According to Proposition~1 and Equation~\ref{eq:T_i}, $\ddbm{v}_i$ is $C^0$-smooth and $T_i$ is at least $C^1$ smooth. When $\gamma_i$ is bounded, $\dot{t}_i$ is also $C^0$-smooth. Hence angular velocity $\boldsymbol{\omega}^\mathcal{I}_i$ is also $C^0$-smooth when $\boldsymbol{f}_{a,i}$ is at least $C^1$-smooth. 
\hfill $\square$


If we use Equation~\ref{eq:angular_rate_reference} to generate the angular velocity reference of each quadrotor, its smoothness is guaranteed through Proposition~2, if we use a smooth drag model (In this work, we assume zero drag for simplicity).
The smooth angular velocity reference, used as feed-forward terms by the trajectory tracking controller onboard the quadrotor, guarantees smooth quadrotor behavior, which is particularly crucial during dynamic motions.

\newpage

\subsection*{Incremental Nonlinear Dynamic Inversion (INDI) Low-Level Controller}
The INDI low-level controller of each quadrotor generates rotor speed commands, using the collective thrust command $T_\mathrm{des}$ and angular acceleration command $\boldsymbol{\alpha}_\mathrm{des}$.
Here, we summarize the key equations of the INDI controller introduced in our previous work~\cite{sun2022comparative}.
In the following context, we denote rotor speed commands as $\boldsymbol{u}_c\in\mathbb{R}^4$, and rotor speed measurement as $\boldsymbol{u}_m\in\mathbb{R}^4$.
Note that the following equations apply to a single quadrotor.
Hence we omit the subscript $i$ for readability. 

The INDI low-level controller employs the following model that maps the rotor speeds to the collective thrust $T\in\mathbb{R}_{\geq 0}$ and body torque $\boldsymbol{\tau}\in\mathbb{R}^3$
\begin{equation}
 \left[\begin{array}{c}
         T  \\
         \boldsymbol{\tau}
    \end{array}\right] = \boldsymbol{G}_1\boldsymbol{u}_m^{\circ 2} + \boldsymbol{G}_2\dot{\boldsymbol{u}}_m
    \label{eq:quadrotor_control_allocation}
\end{equation}
where $\boldsymbol{G}_1$ and $\boldsymbol{G}_2$ are the control effectiveness matrices with respect to the rotor speeds.
Specifically, $\boldsymbol{G}_1$ depends on the shape and size of the quadrotor and aerodynamic coefficients of the propellers.
$\boldsymbol{G}_2$ captures the inertial yawing torque due to the acceleration and deceleration of the rotors, which is a function of the moment of inertia of the rotors. 

Therefore, once the desired collective thrust and torque $\left[T_\mathrm{des},~\boldsymbol{\tau}_\mathrm{des}\right]^\top$ is computed, INDI numerically solves the following equation to obtain rotor speed command $\boldsymbol{u}_c$
\begin{equation}
    \left[\begin{array}{c}
         T_\mathrm{des}  \\
         \boldsymbol{\tau}_{\mathrm{des}} 
    \end{array}\right] = \boldsymbol{G}_1\boldsymbol{u}_c^{\circ 2} + \Delta t ^{-1} \boldsymbol{G}_2\left(\boldsymbol{u}_c - \boldsymbol{u}_{c, k-1}\right)
\end{equation}
where $\Delta t$ is the sampling interval of the controller; $\boldsymbol{u}_{c, k-1}$ is the last computed rotor speed command $\boldsymbol{u}_c$.
And $\boldsymbol{u}_{c, 0} = \boldsymbol{u}_m$ for initialization. 

Unlike conventional dynamic inversion, the INDI low-level controller defines the desired body torque in the following incremental form
\begin{equation}
    \boldsymbol{\tau}_\mathrm{des} = \boldsymbol{\tau}_f + \boldsymbol{J} \left(\boldsymbol{\alpha}_\mathrm{des} - \dot{\boldsymbol{\omega}}_f\right)
\end{equation}
where $\dot{\boldsymbol{\omega}}_f$ is the angular acceleration obtained by numerically differentiating the filtered gyroscope measurement from the quadrotor.
$\boldsymbol{\tau}_f$ is the filtered body torque, which can be calculated using rotor speed measurements leveraging Equation~\ref{eq:quadrotor_control_allocation}, yielding
\begin{equation}
    \boldsymbol{\tau}_f = \left[ \boldsymbol{G}_1\boldsymbol{u}_f^{\circ 2} + \Delta t^{-1}\boldsymbol{G}_2\left(\boldsymbol{u}_f - \boldsymbol{u}_{f,k-1}\right)\right]_{2:4}
\end{equation}
where $\boldsymbol{u}_f$ is the low-pass filtered rotor speed measurements.
Note that the cutoff frequency of the filter for $\boldsymbol{u}_f$ and $\boldsymbol{\omega}_f$ is the same to synchronize the delay introduced by the low-pass filter on these two measurements.
The INDI low-level controller leverages the sensor measurements to effectively capture and compensate for the external torques that are not captured in Equation~\ref{eq:quadrotor_control_allocation}, such as aerodynamic torque, CoG bias of quadrotors, etc.

\newpage
\section*{Supplementary Discussions}
\subsection*{Load Pose Estimation Performance}
\label{sec:pose_estimation}
In the above experiments (all real-world experiments shown in this paper), we need information on load pose and twist to achieve dynamic and accurate trajectory tracking.
In the state-of-the-art method that includes real-world experiments, additional sensors are required for the load pose estimation.
The most commonly used approach is attaching reflective markers on the load to measure its pose from the motion capture system\cite{li2023nonlinear, sanalitro2020full, wahba2024efficient}, or resorting to additional downward-facing cameras and attaching additional circular tags on the load~\cite{li2021cooperative}.
However, it is impractical to attach these sensors in the field for day-to-day operations.

\begin{figure}
    \centering
    \includegraphics[width=1.0\linewidth]{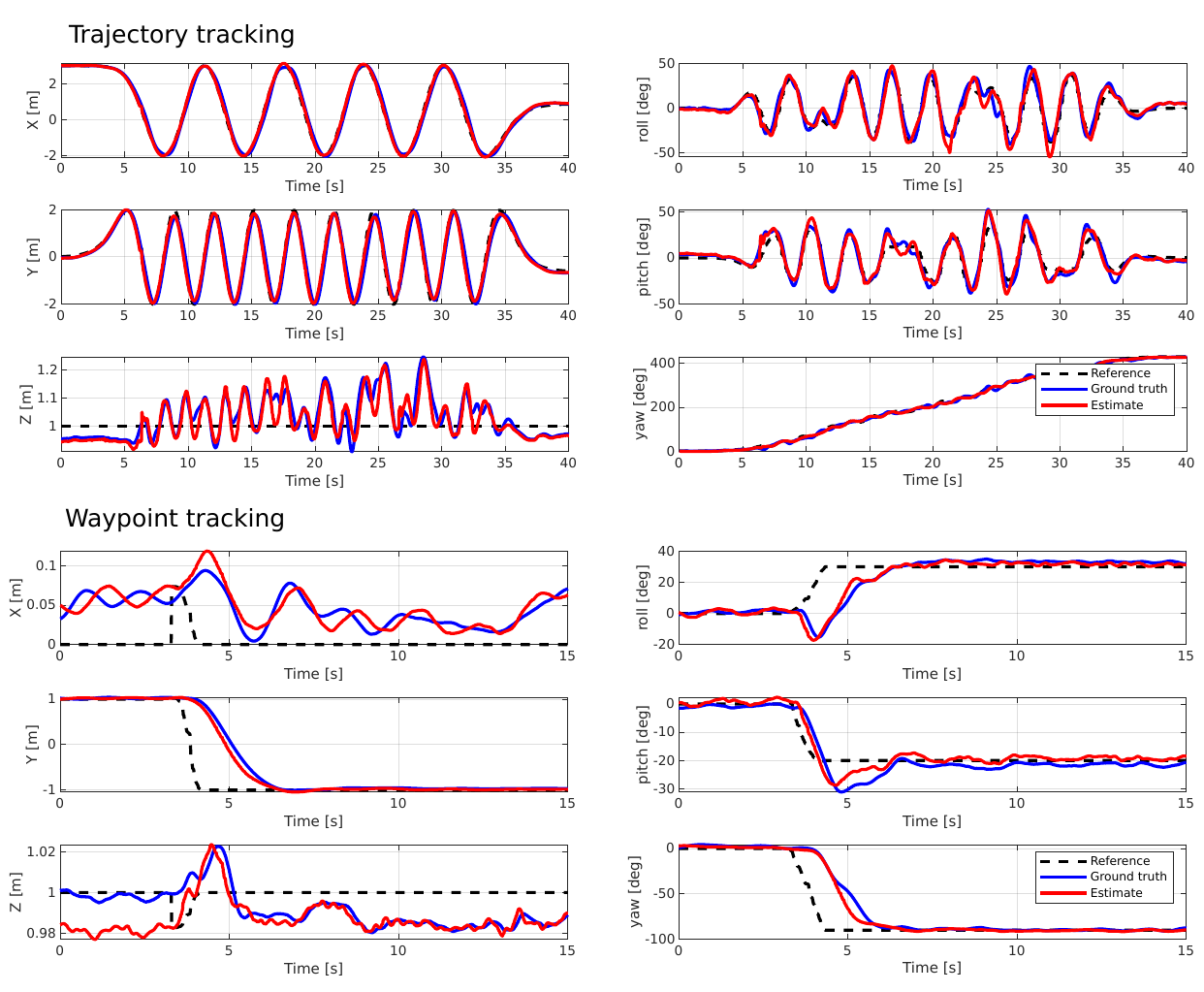}
    \caption{\textbf{Load pose estimation and tracking result.} Without adding additional sensors to the load or force sensors to quadrotors, our algorithm could accurately estimate the load's pose for an accurate closed-loop control. \textbf{Top}: time history of the position and attitude (Euler angles) estimate in comparison with the ground truth from the motion capture system, as well as the reference load pose while tracking the figure-eight trajectory Fast; \textbf{Bottom}: time history of the estimated, the ground truth, and the reference load pose while tracking a setpoint.}
    \label{fig:pose_estimation}
\end{figure}

In comparison, our algorithm does not need to put any additional sensors on the load, nor does it make any modifications to the quadrotors.
In the experiments, we demonstrated that by simply leveraging the IMU on each quadrotor and the dynamic model of the multi-lifting system can provide a sufficiently accurate load pose and twist estimate as well as the cable states to achieve agile pose control. 
Fig.~\ref{fig:pose_estimation} presents the comparison between the ground truth pose of the load and the estimated pose while tracking reference Fast.
Despite the large acceleration of the motion, over 45 degrees of inclination, and continuous yawing motion, our method provided sufficiently high estimation accuracy to achieve closed-loop trajectory tracking.
The position estimation RMSE was ${0.136}~\mathrm{m}$ and the attitude estimation RMSE was ${7.5}~\mathrm{deg}$ even with the highly dynamic motion of the system.

\newpage

\subsection*{Sensitivity to Quadrotor Coordinate Frame Misalignment}
We conducted Monte Carlo simulations to examine the effect of misalignment among ground-fixed reference frames (e.g, odometry frames) used by different quadrotors. 
This issue arises when no global positioning sensors (e.g., motion capture or GPS) are available, such as when using Visual Inertial Odometry (VIO) for state estimation in GPS-denied environments.
In this simulation, we assumed that the initial odometry frames of all quadrotors were well-calibrated and aligned with the inertial frame. 
We then introduced the transformation between the odometry frame and the inertial frame for each quadrotor, in the form of a random-walk process. 
In this way, we simultaneously simulated the misalignment of reference frames among quadrotors and the pose drift typically observed in visual (-inertial) odometry algorithms.

It is worth noting that extensive research has been conducted on aligning estimated reference frames among multiple robots in the context of multi-robot VIO (e.g., \cite{zou2019collaborative, tian2022kimera, cieslewski2021decentralized, peterson2025tcaff}). 
A standard approach is to align these coordinate frames in real-time through place recognition to match landmarks seen by different cameras (quadrotors in our case) and by estimating the relative poses between coordinate frames \cite{cieslewski2021decentralized}.
The accuracy of relative pose estimation depends on the quality frontend / backend in the VIO algorithms and the quality of the established map. 
A typical example alignment frequency is 1~Hz~\cite{peterson2025tcaff}.

Since we were unable to simulate all possible approaches in our study, we only performed a worst-case and best-case scenario analysis. 
In the worst case, we ran simulations without any real-time alignment. 
In the best case, we ran simulations and precisely aligned the quadrotors’ odometry frames with the inertial frame at 1~Hz.
In both cases, we let the system follow the reference trajectory Fast.

\begin{figure}[!h]
    \centering
    \includegraphics[width=1.0\linewidth]{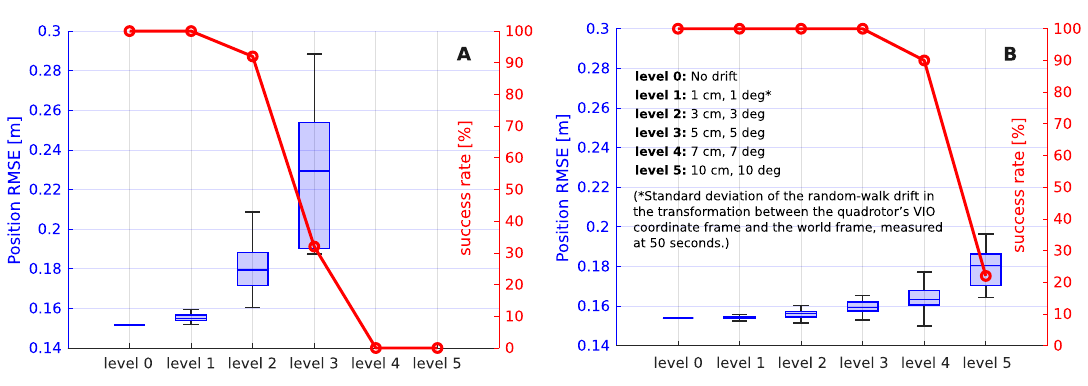}
    \caption{\textbf{Simulation result under misalignment between quadrotor reference frames.} The misalignment was induced by random-walk noise at different levels, on the transform between quadrotor VIO frames and the inertial frame. (\textbf{A}) The worst case: there was no frame alignment mechanism among quadrotors. (\textbf{B}) The best case: the frames were precisely aligned at 1~Hz.}
    \label{fig:rand_walk_noise_misalignment_simulation}
\end{figure}
Fig.~\ref{fig:rand_walk_noise_misalignment_simulation} presents the results of the two cases.
It shows that, in the worst case, without any alignment between quadrotors, our method still maintains a success rate of 90\% at the 2nd noise level, where the standard deviation of coordinate frame drift is 0.03~m on position and 3~deg on attitude after 50 seconds.
We noticed that the attitude misalignment between the coordinate frames is the main cause of failure, as it introduced large position state estimation errors of quadrotors when they follow a trajectory that is far from the origin of the frames.
On the other hand, we demonstrated that with an ideal alignment running at 1~Hz, the performance is almost unaffected by the drift of coordinate frames until it reaches noise level 4.
These results have demonstrated the strong potential of our method to combine with multi-robot VIO algorithms and deploy in a GPS-denied environment.

\clearpage


\begin{figure}
    \centering
    \includegraphics[width=1.0\linewidth]{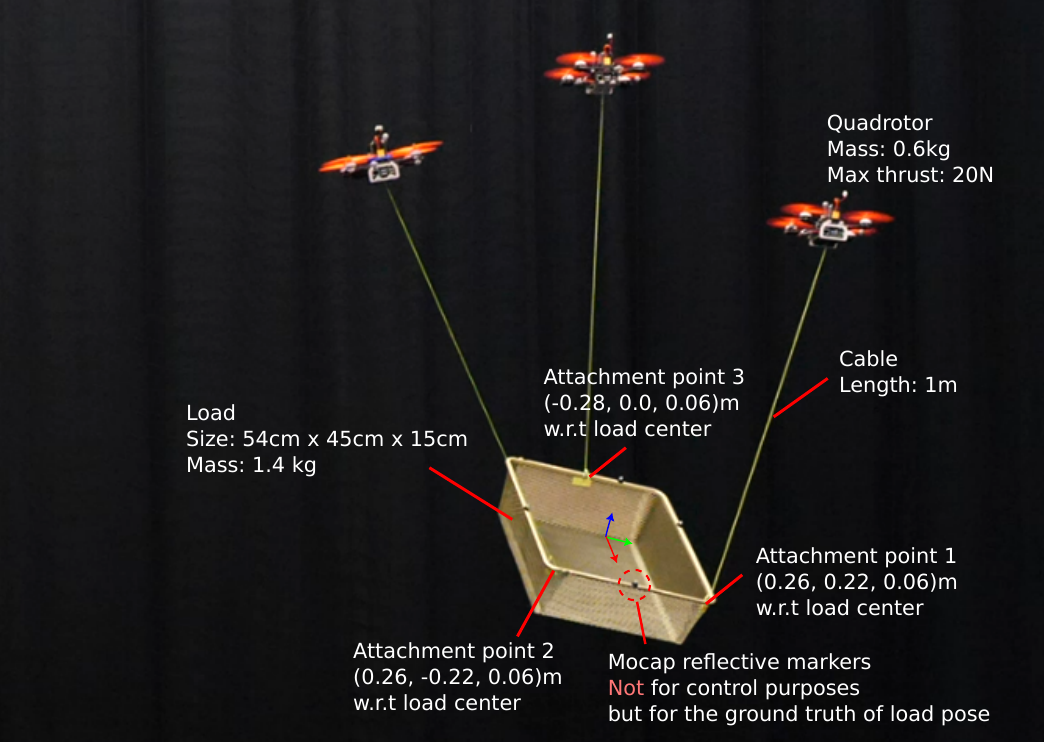}
    \caption{\textbf{Experimental setup.} A snapshot of our experiment, together with the parameters of the load, the quadrotors, and the cables.}
    \label{fig:experimental_setup}
\end{figure}

\clearpage

\begin{figure}
    \centering
    \includegraphics[width=0.8\linewidth]{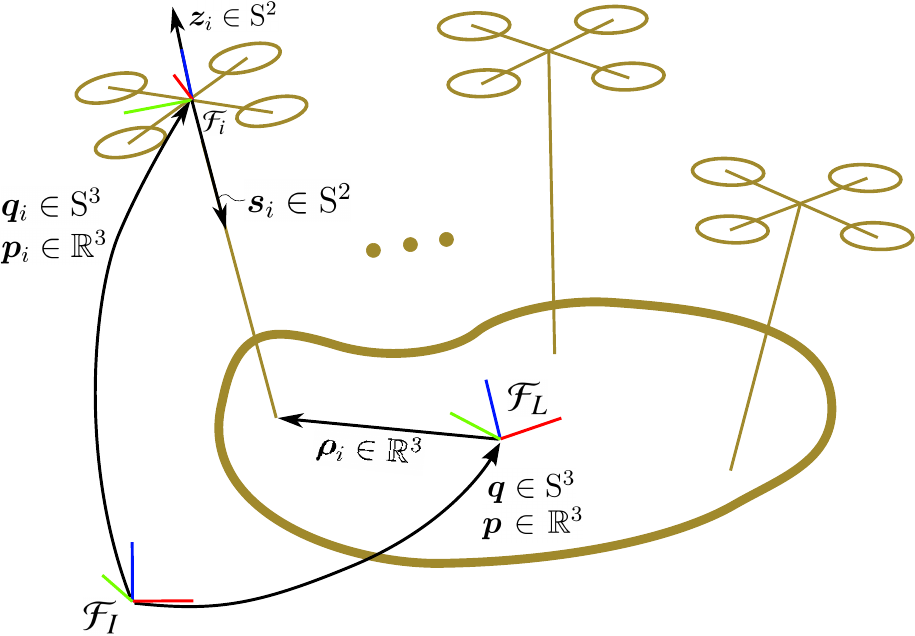}
    \caption{\textbf{Definition of reference frames and symbols.} $\mathcal{F}_I$, $\mathcal{F}_L$, $\mathcal{F}_i$ respectively denote the inertial frame,  load-fixed frame, and the $i$-th quadrotor-fixed frame, where the $x$, $y$, $z$ are marked in red, green, blue colors respectively.}
    \label{fig:symbol}
\end{figure}
\clearpage

\begin{figure}
    \centering
    \includegraphics[width=1.0\linewidth]{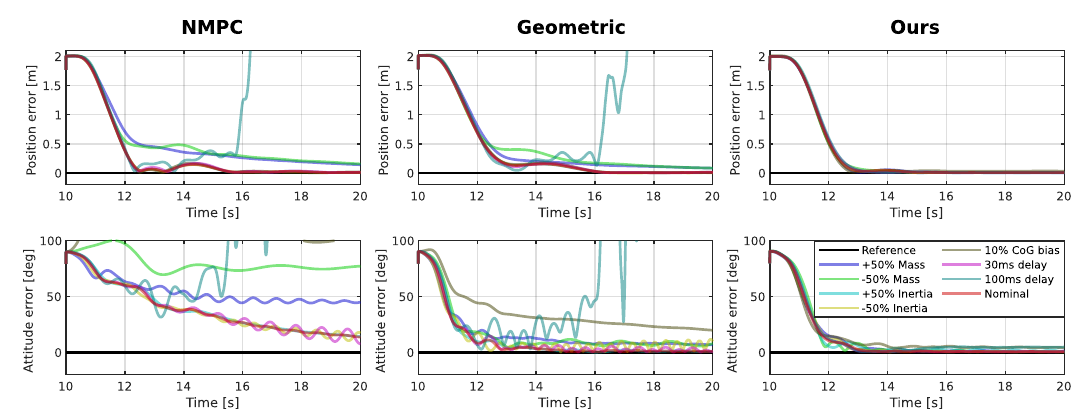}
    \caption{\textbf{Step response under load model uncertainties and communication delay}. Simulation result comparing the tracking performance between our method and the two baseline methods (NMPC\cite{li2023nonlinear} and Geometric~\cite{lee2017geometric}) under various types of model mismatch on the load, as well as communication delay between the centralized planner and quadrotors. We sent a load reference point at 2~m away along y-axis, and an attitude command of -90~deg, -20~deg, and 30~deg, respectively, on yaw, pitch, and roll. Our method clearly outperforms the two baseline methods in the presence of load model mismatch and communication delays.}
    \label{fig:step_response_robustness}
\end{figure}
\clearpage

{
\setlength{\parindent}{0pt}
\setlength{\parskip}{0pt}
\noindent
\parbox{\linewidth}{
  \hrule
  \vspace{0.4ex}
  \textbf{Algorithm S1} \quad Iterative Kabsch–Umeyama algorithm to initialize states of the EKF
  \vspace{0.3ex}
  \hrule
}

\begin{tabbing}
\quad\=\quad\=\quad\=\quad\=\quad\=\kill  
\+
\quad \textbf{Input}: $n,$ $\boldsymbol{p}_i,$ $\boldsymbol{\rho}_i,$ $l_i$ $\text{for } i\in\{1,2,\ldots,n\}$\\
Define tolerance and maximum steps for iteration: $tol_{\text{pos}},\, tol_{\text{att}},\, iter_{\max}$ \\
{Define average cable connection points} $\bar{\boldsymbol{\rho}}=\sum_{i=1}^n\boldsymbol{\rho}_i/n$, $\boldsymbol{L}=[\boldsymbol{\rho}_1-\bar{\boldsymbol{\rho}},~\boldsymbol{\rho}_2-\bar{\boldsymbol{\rho}},...,\boldsymbol{\rho}_n-\bar{\boldsymbol{\rho}}]$ \\
{Define initial load pose} $\boldsymbol{p}=\left[0,~0,~0\right]^\top,\ \ \boldsymbol{R}=\boldsymbol{I}_3$ \\
{Initialize the last load pose} $\boldsymbol{p}_{\text{last}}=\infty,\ \ \boldsymbol{R}_{\text{last}}=\boldsymbol{O}_3$ \\
{Initial guess of cable directions}~$\boldsymbol{s}_i=\left[0,~0,~-1\right]^\top\ \text{for } i\in\{1,2,\ldots,n\}$ \\

\textbf{for } $k=1,\ldots,iter_{\max}$ \textbf{ do} \\
\> \textbf{for } $i=1,\ldots,n$ \textbf{ do} \\
\>\> $\boldsymbol{c}_i = \boldsymbol{p}_i + \boldsymbol{s}_i l_i$ \\
\> $\bar{\boldsymbol{c}}=\sum_{i=1}^n\boldsymbol{c}_i/n$ \\
\> $\boldsymbol{C}=[\,\boldsymbol{c}_1-\bar{\boldsymbol{c}},\ \boldsymbol{c}_2-\bar{\boldsymbol{c}},\ \ldots,\ \boldsymbol{c}_n-\bar{\boldsymbol{c}}\,]$ \\
\> $[\boldsymbol{U}, \boldsymbol{V}] = \textrm{SVD}(\boldsymbol{LC}^{\top})$ \\
\> $\displaystyle
\boldsymbol{R} = \boldsymbol{V}
\left[
\begin{array}{ccc}
1 & 0 & 0\\
0 & 1 & 0\\
0 & 0 & \operatorname{sign}\!\big(\det(\boldsymbol{VU}^{\top})\big)
\end{array}
\right] \boldsymbol{U}^{\top}$ \` $\triangleright$ Estimated load attitude \\

\> \textbf{for } $i=1,2,\ldots,n$ \textbf{ do} \\
\>\> $\tilde{\boldsymbol{p}}_i = \boldsymbol{c}_i - \boldsymbol{R}\boldsymbol{\rho}_i$ \\
\> $\boldsymbol{p}=\sum_{i=1}^{n} \tilde{\boldsymbol{p}}_i/n$ \` $\triangleright$ Estimated load position \\

\> \textbf{for } $i=1,2,\ldots,n$ \textbf{ do} \\
\>\> $\displaystyle
\boldsymbol{s}_i = (\boldsymbol{R}\boldsymbol{\rho}_i + \boldsymbol{p} - \boldsymbol{p}_i) / {\|\,\boldsymbol{R}\boldsymbol{\rho}_i + \boldsymbol{p} - \boldsymbol{p}_i\,\|}$ \` $\triangleright$ Estimated cable direction \\

\> \textbf{if } $\|\boldsymbol{p}-\boldsymbol{p}_{\text{last}}\|<tol_{\text{pos}} \ \textbf{and}\ \|\det(\boldsymbol{R}^{-1}\boldsymbol{R}_{\text{last}})-1\|<tol_{\text{att}}$ \textbf{ then} \\
\>\> \textbf{Break} \\

\> $\boldsymbol{p}_{\text{last}}=\boldsymbol{p},\ \ \boldsymbol{R}_{\text{last}}=\boldsymbol{R}$ \\

\textbf{Return} $\boldsymbol{p},\ \boldsymbol{q}(\boldsymbol{R}),\ \boldsymbol{s}_1,\boldsymbol{s}_2,\ldots,\boldsymbol{s}_n$ \\
\end{tabbing}
\hrule

}

\clearpage
\begin{table}
\small
\centering
\caption{\textbf{Algebraic expression of load reference trajectories}}
\begin{tabular}{l|ccc}
\hline
Name & $\boldsymbol{p}_x$& $\boldsymbol{p}_y$& $\boldsymbol{p}_z$\\ \hline
Slow       & $2.5\cos{(0.25 t)}$ & $2\sin{(0.5 t)}$ & $1.0$
\\
Medium     & $2.5\cos{(0.5 t)}$ & $2\sin{(t)}$ & $1.0$ \\
Medium Plus     & $\cos{(t)}$ & $\sin{(2t)}$ & $1.0$\\
Fast      & $2.5\cos{(t)}$ & $2\sin{(2t)}$ & $1.0$ 
\\ \hline
\end{tabular}
\label{tab: algebratic_load_reference}
\end{table}

\clearpage 

\paragraph{Caption for Movie S1.}
\textbf{Video comparing our method with baseline methods in simulation}. The video shows the 6-DoF simulation result of our method and the two baseline methods in tracking the two reference trajectories (Fast and Slow). The video is associated with the result given in Table~\ref{tab: trajectry_tracking_comparison}.

\paragraph{Caption for Movie S2.}
\textbf{Video of flight for obstacle avoidance}. The video shows real-world experiments performing two obstacle avoidance tasks: one through a narrow passage between two walls, and the other through a horizontally oriented gap.

\paragraph{Caption for Movie S3.}
\textbf{Video of flight with four quadrotors}. The video shows the real-world experiment of our method scaling up to a case with four quadrotors following a reference trajectory dynamically.

\paragraph{Caption for Movie S4.}
\textbf{Video of flight in windy conditions}. The video shows that our method effectively controls the system to follow trajectories under moderate wind disturbances.

\paragraph{Caption for Movie S5.}
\textbf{Video showing results with large quadrotor state estimation errors}. The video shows real-world experiment and simulation results under different levels of quadrotor state estimation errors.




\end{document}